\documentclass{article}

\usepackage{PRIMEarxiv}
\usepackage[table, xcdraw]{xcolor}

\usepackage[utf8]{inputenc} 
\usepackage[T1]{fontenc}    
\usepackage{hyperref}       
\usepackage{url}            
\usepackage{booktabs}       
\usepackage{amsfonts}       
\usepackage{nicefrac}       
\usepackage{microtype}      
\usepackage{lipsum}
\usepackage{fancyhdr}       
\usepackage{graphicx}       
\usepackage{todonotes}
\usepackage{multicol}

\usepackage{subcaption}
\graphicspath{{media/}}     
\usepackage{amsmath,amssymb}

\usepackage[section]{placeins}

\title{Gradient-free neural topology optimization: towards effective fracture-resistant designs}

\author{
  Gawel Kus \\
  School of Engineering \\
  Brown University \\
  Providence, RI\\
   \And
  Miguel A. Bessa \\
  School of Engineering \\
  Brown University \\
  Providence, RI\\
  \texttt{miguel\_bessa@brown.edu} \\
}

\begin{document}

\maketitle

\begin{abstract}
Gradient-free optimizers allow for tackling problems regardless of the smoothness or differentiability of their objective function, but they require many more iterations to converge when compared to gradient-based algorithms. This has made them unviable for topology optimization due to the high computational cost per iteration and the high dimensionality of these problems. We propose a gradient-free neural topology optimization method using a pre-trained neural reparameterization strategy that addresses two key challenges in the literature. First, the method leads to at least one order of magnitude decrease in iteration count to reach minimum compliance when optimizing designs in latent space, as opposed to the conventional gradient-free approach without latent parameterization. This helps to bridge the large performance gap between gradient-free and gradient-based topology optimization for smooth and differentiable problems like compliance optimization, as demonstrated via extensive computational experiments in- and out-of-distribution with the training data. Second, we also show that the proposed method can optimize toughness of a structure undergoing brittle fracture more effectively than a traditional gradient-based optimizer, delivering an objective improvement in the order of 30\% for all tested configurations. Although gradient-based topology optimization is more efficient for problems that are differentiable and well-behaved, such as compliance optimization, we believe that this work opens up a new path for problems where gradient-based algorithms have limitations.
\end{abstract}


\maketitle

\section{Introduction}
\label{sec:intro}
Topology optimization plays an important role in the design of structures \cite{sigmund2013topology,munk2019benefits, zhu2016topology}. Historically, interest in gradient-free optimization has been present since the early days of the topology optimization field \cite{papadrakakis1998structural, xie1993simple}, motivated mostly by their ability to deal with discrete design variables and their generality. However, with the advancements of the continuous design variable formulation, the developments focused on linear and differentiable problems, such as compliance minimization -- suitable for a more efficient gradient-based optimization \cite{sigmund2013topology, woldseth2022use, wu2021topology,sigmund200199}. While these problems are of high academic importance, many practical scenarios are governed by non-differentiable, non-linear objectives that are much more challenging to optimize for \cite{desai2022topology, da2018topology, schwarz2001topology,guirguis2019evolutionary}. In principle, gradient-free optimization algorithms could be a promising approach for solving such problems because they can be applied to any objective, thus remaining a relevant alternative. However, they suffer from a major limitation: they require several orders of magnitude more objective evaluations (simulations) than gradient-based optimizers to converge to a solution \cite{sigmund2011usefulness,guirguis2019evolutionary}.

Gradient-free optimizers update the solution by sampling and comparing the performance of trial solutions. At each iteration, the objective needs to be evaluated for the whole population of samples, using expensive simulations (usually FEM). Moreover, gradient-free optimizers suffer from the curse of dimensionality, i.e. the cost of optimization grows exponentially with the number of design variables \cite{hansen2010comparing, sigmund2011usefulness}. This is especially nefarious to topology optimization, as the number of design variables is typically very large, rendering these optimizers unfeasible \cite{sigmund2013topology, sigmund2011usefulness}. 

Taking into account these limitations, it comes as no surprise that gradient-free optimization is not well motivated for most topology optimization problems \cite{sigmund2011usefulness}. However, there are scenarios where the use of gradient-based methods might be challenging, namely in cases where the gradients are difficult to compute or where the objective function is noisy or discontinuous \cite{desai2022topology, russ2019topology, guirguis2019evolutionary,huang2023topology}. In principle, gradient-free methods would be a logical solution in these contexts, but their slow convergence detracts from their use \cite{sigmund2011usefulness,guirguis2019evolutionary}.

In this article, we address the large rift in performance between gradient-free and gradient-based topology optimization for problems where the latter is orders of magnitude more efficient than the former. Therefore, we consider thousands of topology design problems where gradient-based optimizers severely outperform the best gradient-free algorithms and show that an appropriate machine learning strategy combining training (offline) and neural reparameterization (online) closes the performance gap by an order of magnitude. 

Compared to the state-of-the-art latent optimization of topology with gradient-free methods \cite{guo2018indirect}, our work explicitly targets the issue of scalability of gradient-free optimization. We show that the appropriate choice of the model architecture leads to significant gains in performance, compared to the conventional approach (without latent space reparameterization), but also compared to more standard architectures. We test our approach on out-of-distribution examples, i.e. examples that were not used in training and that include characteristics that were not explored in training. 

We demonstrate the effectiveness of our proposed approach by deploying it on a non-linear path-dependent optimization problem that proves to be challenging for standard gradient-based approaches -- namely optimizing structures undergoing brittle fracture. We test different initialization strategies and show that the gradient-based optimization remains prone to getting stuck in local minima, while the gradient-free optimizer coupled with latent reparameterization delivers improved performance without retraining or tuning the hyperparameters. Notwithstanding, we do not claim that the proposed approach can outperform gradient-based methods in convex or nearly convex objective functions, as occurs in compliance optimization problems. However, we believe that this research is a first step towards solving more challenging problems in the future, involving plasticity, fracture, and fatigue.

\section{Related work}

\subsection{Limitations of gradient-based and gradient-free approaches}

In many optimization problems, the objective function is differentiable, such as optimization of compliance, and the derivatives can be obtained analytically or with automatic differentiation \cite{88line, hoyer2019neural, sigmund2013topology, zhang2021tonr}. In case the function is nondifferentiable, so far, the most widespread approach is to linearize the problem such that it can still be solved with a gradient-based optimizer, in particular using the adjoint method \cite{jameson2007adjoint, schwarz2001topology, da2018topology, giraldo2020unified, zhao2019material, zhang2020topology}. Although relatively successful, this approach comes with certain limitations, as highlighted for example in \cite{desai2022topology}, where the authors use the adjoint formulation of the topology optimization problem to postpone fracture. The presented derivation of the adjoint assumes that the objective function is smooth, however, this is not the case for fracture, where even the smallest changes in the design can have a critical impact on the performance (specifically, these changes can completely change the crack trajectory). As highlighted by the authors, the effects cannot be easily remedied by fixing the step size of the descent algorithm or by regularization. In such a case, when the objective function is not well behaved, discontinuous, with multiple local minima, and difficult to regularize, gradient-free methods could be a promising alternative \cite{sigmund2011usefulness, desai2022topology,huang2023topology,guirguis2019evolutionary}.

Several gradient-free algorithms were explored in topology optimization, including Genetic Algorithms \cite{balamurugan2008performance,balamurugan2011two, ramamoorthy2023multi}, Artificial Immune Algorithms \cite{luh2004multi}, Ant Colonies \cite{kaveh2008structural,luh2009structural}, Particle Swarms \cite{luh2011binary, plevris2011hybrid}, Simulated Annealing \cite{shim1997generating}, Harmony Search \cite{lee2004new}, Differential Evolution \cite{wu2010topology} among others \cite{sigmund2011usefulness, guirguis2019evolutionary}. These approaches, however, received strong criticism in the community \cite{sigmund2011usefulness} primarily due to their inferior efficiency, as opposed to gradient-based methods. Gradient-free algorithms suffer from the curse of dimensionality \cite{sigmund2011usefulness, hansen2010comparing}, which means that the cost of optimization increases approximately exponentially with the number of problem dimensions. In standard topology optimization approaches \cite{88line}, where the design variables correspond to element-wise densities, the dimensionality of even small 'toy-examples' is in practice too large for gradient-free algorithms, leading to prohibitive costs (in order of 10,000 - 100,000 objective function evaluations \cite{plevris2011hybrid}, as opposed to the order of 10 -- 100 evaluations for gradient-based methods \cite{sigmund2011usefulness}).

\subsection{Latent-space optimization}
\label{sec:headings}

The curse of dimensionality of gradient-free optimizers can be addressed by reparameterizing the optimization problem into a lower-dimensional latent space \cite{guirguis2019evolutionary,lu2018structured, notin2021improving, tripp2020sample}. This dimensionality reduction can be well accommodated using generative deep learning models, such as generative adversarial networks (GAN) \cite{goodfellow2014generative}, variational autoencoders (VAE) \cite{kingma2013auto}, and diffusion models \cite{sohl2015deep, dhariwal2021diffusion}. In the context of latent space optimization, variational autoencoders are a particularly common choice \cite{park2022optimization, griffiths2020constrained}. 

Optimizing in the latent space of a VAE was demonstrated in the context of topology optimization \cite{sato2023fast, gladstone2021robust, guo2018indirect, schumann2021machine}. Sato et al. \cite{sato2023fast} and Gladstone et al. \cite{gladstone2021robust} explored latent optimization with a gradient-based approach using a surrogate model -- a neural network trained to predict the property of interest from the latent representation. The method takes advantage of the differentiability of the surrogate -- as the property of interest can be minimized by backpropagating through the surrogate. In both cases, the results demonstrate that the VAE is capable of representing new designs, that outperform those seen in the training. The disadvantage of this approach, however, is the need for data, required for training the surrogate. 

Guo et al. \cite{guo2018indirect} explore topology optimization in the latent space of a VAE with different optimization algorithms, including a gradient-free genetic algorithm. Unlike other works, the training of the VAE was augmented with an additional style-transfer procedure, to improve the quality of the generated designs. The training dataset comprised designs optimized for thermal compliance using the SIMP (solid isotropic material with penalization) method \cite{88line}, but the model was then used to solve a different problem - a multi-objective optimization, minimizing the maximum temperature and maximizing the power density. Similar to other works \cite{sato2023fast,gladstone2021robust}, the method was able to synthesize novel designs, different from those seen in the training dataset. Most importantly, the results demonstrate that the pre-trained model can be successfully deployed on a different optimization problem, given the designs have similar features. Nevertheless, in the presented experiments the optimization process with the gradient-free method (Genetic Algorithm) required an order of 30,000 FEM evaluations, indicating relatively limited improvement as compared to the gradient-free optimization without reparameterization into the latent space \cite{sigmund2011usefulness}.     

\subsection{Machine learning in topology optimization}

In the context of topology optimization, generative machine learning models have been primarily used to map boundary conditions to final designs \cite{shin2023topology, woldseth2022use, ramu2022survey}. Different generative models were explored in such a setup, including (GANs) \cite{nie2021topologygan, oh2019deep} and diffusion models \cite{maze2023diffusion}. While demonstrating competitive performance, these approaches are often criticized for their limited generalization \cite{shin2023topology, woldseth2022use}, as they remain applicable only to the problem on which they were trained. In order to apply such a model to a new type of design problem (e.g. different objective, different physics), the model needs to be retrained on a dataset corresponding to the new problem. Generating the dataset, however, requires the use of conventional optimization methods and simulations. 

Another relevant stream of work is neural reparameterization \cite{hoyer2019neural, chandrasekhar2021tounn, zhang2021tonr, zhang2023topology,chandrasekhar2021length, woldseth2022use,joglekar2023dmf,zhong2022nsto,sanu2024neural,yousef2024}, where the neural network is used to implicitly represent the designed structure. Instead of optimizing the element-wise density values, the optimizer adjusts the trainable parameters of a neural network which outputs the density field. This approach proved beneficial in gradient-based approaches, where the weights can be adjusted by back-propagating the gradients of the density field (design sensitivities) \cite{hoyer2019neural}. Neural reparameterization was not explored in a gradient-free optimization scenario. One reason is the relatively high number of parameters of the neural networks, ranging from 100s \cite{chandrasekhar2021length} to 100,000s \cite{hansen2001completely}, making most network architectures unfeasible to optimize with gradient-free algorithms. 

\section{Methods}

\begin{figure*}
\centering
\includegraphics[width = \textwidth]{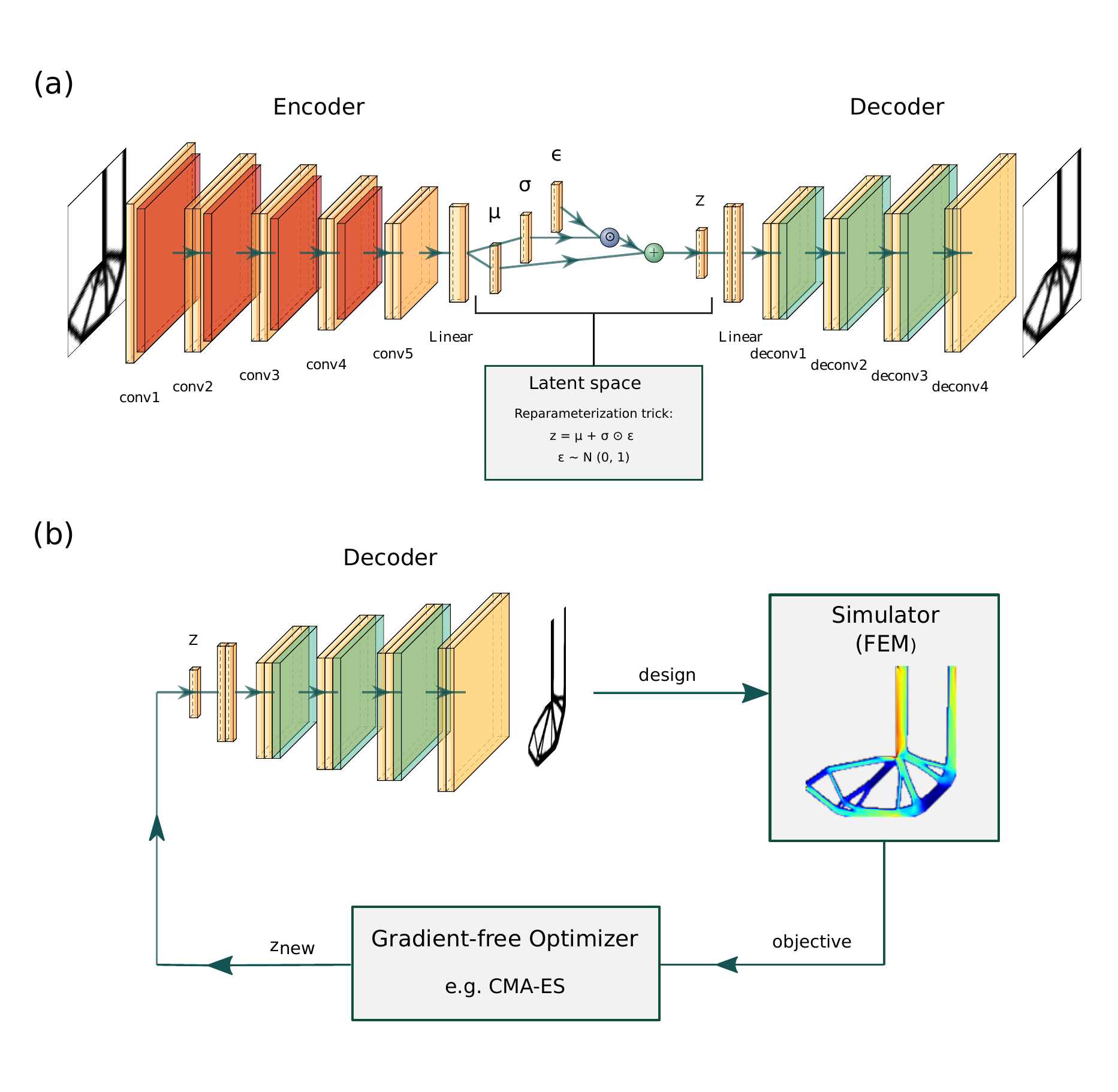}
     \caption{Schematics of our framework. a) Schematic of a variational autoencoder (VAE). In the first step of our method, we train the generative model -- a VAE to reparameterize topology designs into latent space. Here $\mu$ and $\sigma$ are mean and variance vectors in latent space, $\epsilon$ is the noise vector sampled from a multivariate Gaussian, and $z$ is a latent variable. b) Schematic of latent space optimization process with gradient-free optimizer. In the second step of our method, we use a trained generative model, to optimize the designs using latent space representation of the generative model. Note that the latent vector $z$ is no longer stochastic, and it is fully controlled and updated ($z_{new}$) by the optimizer, based on the objective values obtained from the simulator.}
     \label{fig:schematic_all}
\end{figure*}

We propose a latent space optimization approach following two steps, as shown on the schematic in Figure \ref{fig:schematic_all}. First, we train a generative model to reparameterize the topology designs into a lower-dimensional latent representation. Then, in the second step, the latent space is explored by the gradient-free optimizer, which attempts to find a design minimizing the objective. The optimizer controls the latent vector which parameterizes the design. This vector is decoded into a physical design using the pre-trained generative model, and the design is evaluated using the objective function (e.g. compliance). During one iteration of optimization, the optimizer generates a population of trial latent vectors, which are subsequently evaluated, and based on their associated objective values, the optimizer updates the population according to its internal update rules, converging to a better solution. 

The key assumption of our approach is that the generative model learns compact representations of features occurring in structural designs, that can be shared between designs optimized for different objectives (e.g. compliance, minimum mass, or compliant mechanism design). Evidence of this can be found in \cite{desai2022topology} or \cite{russ2019topology}, by comparing designs optimized for fracture with those optimized for standard compliance. This observation can be verified using quantitative feature similarity metrics such as Learned Perceptual Image Patch similarity \cite{zhang2018unreasonable} (for details, see the appendix). Based on that, we assume that the generator trained to capture visual features can be deployed to optimize different objectives without the need for retraining and is not strictly limited to the problems used for generating the training dataset. This is a crucial advantage over state-of-the-art approaches exploring generative models for topology optimization.

\subsection{Generative Model}

We propose to use a new VAE variant called Latent Bernoulli Autoencoder (LBAE) that uses the Bernoulli distribution for modeling the latent space \cite{pmlr-v119-fajtl20a}. Although this architecture was proposed in a different context, it has been shown to achieve superior performance compared to conventional VAE architectures in terms of the quality of the generated samples (FID score) on a number of image generation tasks \cite{pmlr-v119-fajtl20a}. We conjectured that using a Bernoulli distribution for the latent space (zeros and ones), instead of a continuous distribution such as a Gaussian, is particularly adequate for the case of topology optimization where the aim is to decide whether each discrete element should contain material or not. We carefully compare the LBAE to a conventional VAE architecture and search for the best hyperparameters for a large class of problems, in an attempt to consider each method at their best performance \cite{guo2018indirect}. Furthermore, we include gradient-based results and different gradient-free optimizers to establish clear performance metrics and clarify the usefulness of the proposed method.

Conventionally, VAE provides a generative model that captures the distribution of data \(\mathcal{D} = \left\{ \mathbf{x}^{(1)}, \mathbf{x}^{(2)}, ..., \mathbf{x}^{(N)} \right\}\) assuming a joint distribution $p(\mathbf{x}, \mathbf{z})$ using some hidden (latent) variable \(\mathbf{z}\), such that given \(\mathbf{z}\), it can generate samples of \(\mathbf{x}\). In practice, this is achieved by using two neural networks: the encoder, which approximates the posterior \(p(\mathbf{z} | \mathbf{x})  \approx q_\phi (\mathbf{z}|\mathbf{x})\), (since in general \(p(\mathbf{z} |\mathbf{ x}) \) is intractable), and the decoder, which captures \(p_{\mathbf{\theta}}(\mathbf{x}|\mathbf{z})\). The two networks are trained simultaneously by maximizing the evidence lower bound, as shown in Equation (\ref{eq:loss}). 
\begin{align}
    & L_{\mathbf{\theta}, \mathbf{\phi}} = 
    \mathbb{E}_{q_{\mathbf{\phi}}
    (\mathbf{z}|\mathbf{x})}
    \left[ \log \frac{p_{\mathbf{\theta}}(\mathbf{x}|\mathbf{z})p(\mathbf{z})}{q_{\mathbf{\phi}}(\mathbf{z}|\mathbf{x})}  \right]  \nonumber \\
    &= \mathbb{E}_{q(\mathbf{z}|\mathbf{x})}\left[\log p_{\mathbf{\theta}}(\mathbf{x}|\mathbf{z})\right] - D_{KL}\left( q_{\mathbf{\phi}}(\mathbf{z}|\mathbf{x})||p(\mathbf{z}) \right)
    \label{eq:loss}
\end{align}
The first term in Equation (\ref{eq:loss}) corresponds to the reconstruction loss, and the second term corresponds to the Kullback-Leibler divergence, which regularizes the latent variable and aims to drive the parameterized distribution towards the assumed prior. In a conventional VAE, the prior $p(\mathbf{z})$ is assumed to be a multivariate Gaussian, while in LBAE the latent space is assumed to be a multivariate Bernoulli distribution. Unlike in VAE, however, in LBAE the Bernoulli distribution of the latent variable is imposed via network architecture and its specific sampling routine, rather than by an explicit regularization term (for details see \cite{pmlr-v119-fajtl20a}). Therefore, the second term of Equation (\ref{eq:loss}) is not present in the LBAE loss. 

The training dataset $x \sim X$ in our case corresponds to images representing different topology designs. Once trained, the autoencoder is then used to generate new designs by sampling $z$ from the latent space. In this way, the designs can be represented using a lower-dimensional latent vector.

\paragraph{Architecture}

The LBAE architecture that we use is based on the model proposed by Fajtl et al. \cite{pmlr-v119-fajtl20a}, adapted to our dataset by adjusting the number of layers and convolutional filters. According to the authors \cite{pmlr-v119-fajtl20a}, modeling latent space with Bernoulli distribution produces sharper images, compared to traditional VAE, and allows for smoother interpolation in the latent space. We hypothesize that these properties are crucial for optimization, allowing for easier traversing of the loss landscape by the optimizer. Furthermore, unlike in the baseline VAE, LBAE uses residual layers (for details, see Appendix \ref{app:E}, Tables \ref{tab:lbae-encoder} and \ref{tab:lbae-decoder}). Combined with the Bernoulli distribution of the latent variable, we observed that this architecture is indeed capable of producing output with finer detail and reduced blurriness, addressing one of the main shortcomings of VAE \cite{woldseth2022use}. The baseline VAE architecture used in our case is a modified model of Larsen et al. \cite{larsen2016autoencoding}. The inputs and outputs are single-channel, grayscale representations of the designs of resolution 64x64.

For both LBAE and VAE, one of the most important parameters is the dimensionality of the latent space, which balances the training and the optimization performance. While increasing the latent space dimensionality makes the training easier, as the compression of information is not as severe, the optimization becomes less efficient due to the curse of dimensionality \cite{sigmund2011usefulness, hansen2010comparing}. To find a balanced latent space dimensionality, we tested several configurations (for details, see Appendix \ref{app:d}). In the case of VAE, we did not observe a significant difference between the dimensionality 32 and 64, instead, we noticed that the performance was limited by insufficient resolution of finer details in the output (caused by the blurriness intrinsic to the VAE). These limiting factors, however, were to a large extent alleviated in LBAE, where the trade-off due to the latent space dimensionality was much more clear (for details, see the appendix). We found that for LBAE the dimensionality 256 offered the best balance. 

\paragraph{Dataset}

We use two different datasets of topology designs at train and test time, respectively. The training dataset was generated based on a modified approach of \cite{sosnovik2019neural} and \cite{maze2023diffusion}, resulting in 48,872 designs.  The boundary conditions were generated as randomly drawn point loads and point supports, using Poisson's distribution specified in the same way as in \cite{sosnovik2019neural}. Additionally, in our dataset, we randomly applied a square mask to restrict part of the design domain from applying material (e.g., for an L-shaped beam, the mask would be applied in the corner). The mask was applied with 25\% probability, and the position of the mask was randomly assigned along one of the edges of the design domain. Volume fractions were drawn using a uniform distribution with a range of [0.12, 0.5]. This dataset was split into two subsets: for training and validation, using a standard 80:20 split.

The testing dataset was used only for monitoring the testing loss, which aims to verify the generalization capability of the model. This dataset was generated using the problems defined in \cite{hoyer2019neural}. The boundary conditions were generated using one of 28 classes of parameterized problems, with the parameters drawn at random from a uniform distribution, which resulted in over 4000 instances of testing problems. Unlike the training problem set, the testing problem classes include distributed loads and supports; therefore, they are expected to provide a good insight into the generalization of the model and can be considered out-of-distribution.

The designs in both datasets were obtained using a gradient-based MMA (Method of Moving Asymptotes) \cite{svanberg1987method} optimizer with the default hyperparameter settings. The final designs were represented as grayscale images of fixed resolution (64x64) with pixel values representing the volume fraction of the material. For each problem, the optimization with MMA \cite{svanberg1987method} was run for 200 iterations. The simulations were set up following the approach of the 88-line code \cite{88line}, with $E_0 = 1.0$, $E_{min} = 1e-9$, $\nu = 0.3$ and a standard cone filter with a radius of 2 pixels to regularize the density field.  

\subsection{Benchmarking set-up}

We develop and benchmark our method considering a standard topology optimization problem, namely compliance minimization with a volume constraint. This problem can be easily solved using gradient-based approaches, so it is not the ultimate application of the proposed method. Nevertheless, it provides a good assessment for the development of the method because, first, the ``ground truth'' solutions can be easily obtained using a gradient-based approach, allowing us to verify 'how far' our method is from a good solution. Secondly, these problems are relatively cheap to evaluate, as opposed to design problems based on more complex physics, such as the fracture problem presented towards the end of the article. 

\paragraph{Benchmarking problem sets}

To provide statistically meaningful performance measures, we test our optimization framework on two sets of optimization problems. The first set contains 25 in-distribution design problems (sets of boundary conditions). These problems were generated in the same way as the problems seen during the training (i.e., with random point loads and point supports). The second set contains 25 out-of-distribution problems, generated using the approach of Hoyer et al. \cite{hoyer2019neural} (the same as the testing dataset). These problems include distributed loads and supports, therefore they provide a better insight into the generalization of the model.

\paragraph{Performance metrics}

To quantify the performance of different model and optimizer configurations across the whole problem set, for each problem, we normalize the compliance value with the compliance of the MMA solution (considered as the 'ground truth'), and in this way calculate the relative error. Furthermore, to account for stochastic factors within the optimization process (e.g., initialization of the optimizer), the optimization process for each configuration and each problem is repeated with 15 different random seeds. Performance is evaluated for grayscale and binary (thresholded) designs. For thresholding we apply the simplest strategy from \cite{sigmund2022benchmarking} in a post-processing step. In that case, the relative error is calculated with respect to the value obtained for the thresholded MMA design.

\paragraph{Gradient-free optimizer}

For gradient-free optimizer, we use Covariance Matrix Adaptation Evolutionary Strategies (CMA-ES) \cite{hansen2001completely, evosax2022github} and its Bi-Population variant (BIPOP-CMA-ES) \cite{hansen2009benchmarking}. The CMA-ES and its variants (such as BIPOP-CMA-ES) are currently regarded as state-of-the-art gradient-free optimization algorithms \cite{hansen2010comparing}. Furthermore, CMA-ES was considered in a similar setting \cite{on2021optimal}. Notwithstanding, we considered several alternative gradient-free optimizers to ensure that we selected the most favorable one for our problems, namely Particle Swarm Optimization \cite{kennedy1995particle}, Differential Evolution \cite{storn1995differrential}, Evolutionary Strategies meta-optimizer \cite{lange2023discovering}, and Simple Genetic Algorithm \cite{such2017deep}. However, even with hyperparameter tuning, these optimizers exhibited considerably worse performance than CMA-ES within the assigned budget. 

\paragraph{Topology optimization set-up}
We use the density-based formulation of topology optimization, which is the most widely adopted approach, although numerous other formulations are available \cite{sigmund2013topology}. The volume constraint is enforced following the approach of Hoyer et al. \cite{hoyer2019neural}, by mapping the outputs of the neural network $\tilde{x}$ to density values $x$ with the following sigmoid transformation:
\begin{align}
x = \frac{1}{1  + \exp{\left( \tilde{x} - b(\tilde{x}, V_0)\right)}}  && s.t: \int_{\Omega}\rho = V_0
\end{align}

where the constant $b(x, V_0)$ is found using a bisection algorithm, such that the final volume of the density field over the design domain $\Omega$ satisfies the prescribed volume $V_0$.

For the remaining parts, we use the same set-up as for the MMA baselines - i.e. the cone filter with a radius of 2 pixels, and the material parameters set to $E_0=1.0$, $E_{min} = 1e-9$ and $\nu = 0.3$.

\section{Results}

\begin{figure*}[h]
      \includegraphics[width=\textwidth]{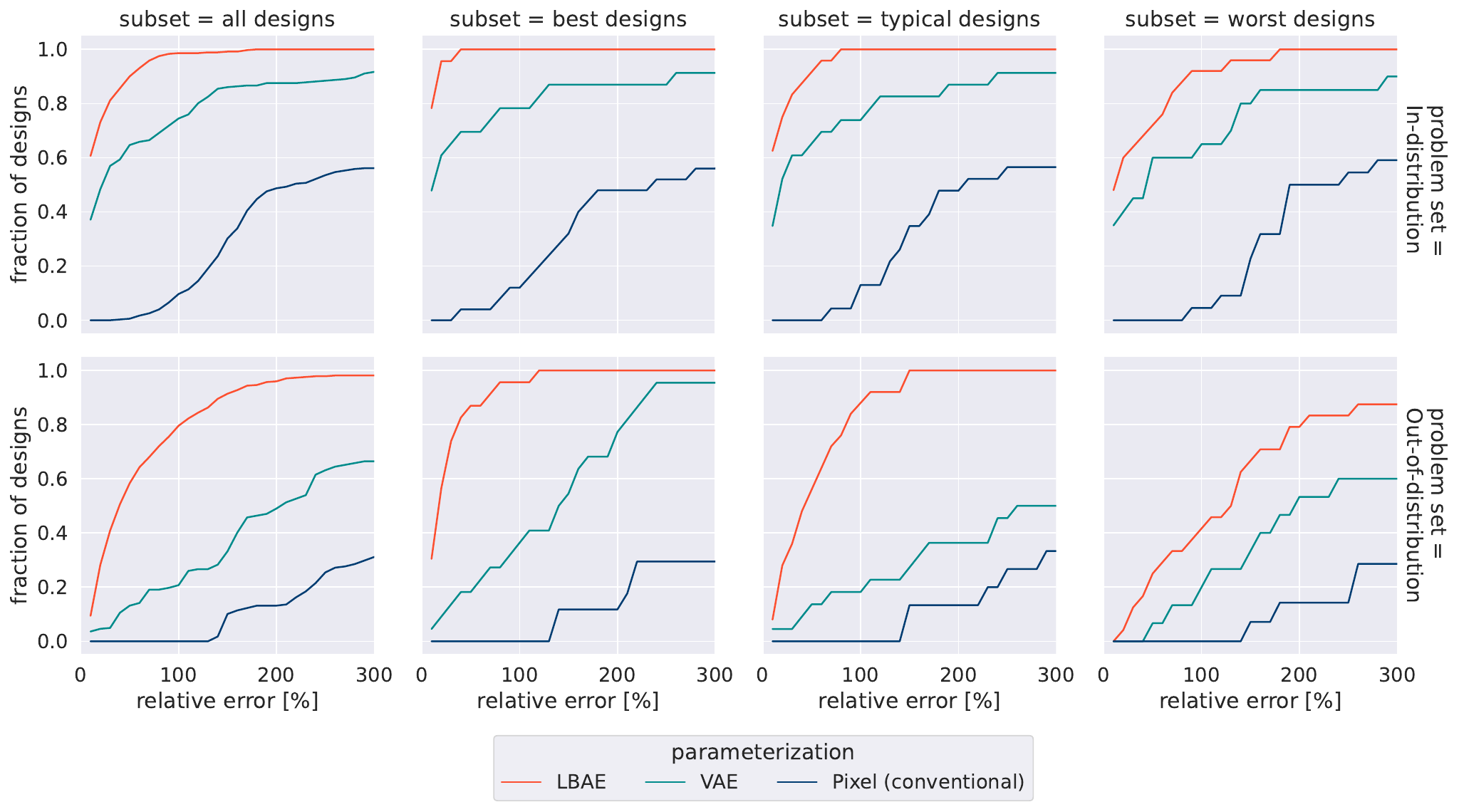}
      \caption{Performance of unprocessed designs: cumulative probability (fraction of problems) vs relative error w.r.t. the MMA solution, for two variants of latent parameterization models -- Variational Autoencoder (VAE, latent space dimensionality: 64), and Latent Bernoulli Autoencoder (LBAE, latent space dimensionality: 256) -- compared against conventional pixel parameterization. The designs were optimized with a BIPOP-CMA-ES optimizer. Note that the shown distributions do not add up to 1 due to the cut-off at 300\%. Best, typical, and worst designs correspond to the best, median, and worst random initializations for each problem.}
    \label{fig:stats1}
\end{figure*}

\begin{figure*}
  \centering
  \includegraphics[width=\textwidth]{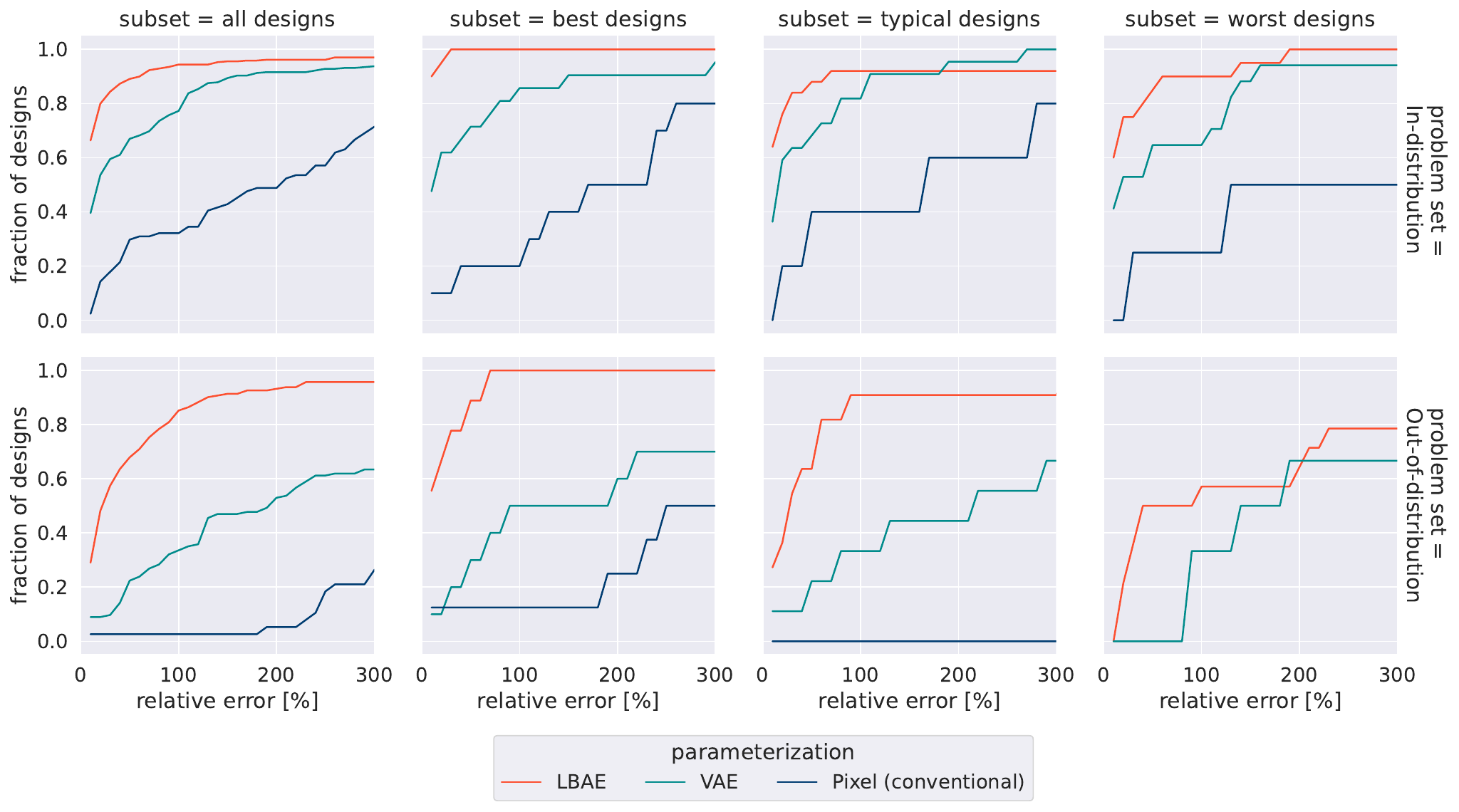}
  \caption{Performance of thresholded designs: cumulative probability (fraction of problems) vs relative error w.r.t. the MMA solution, calculated for thresholded designs for two variants of latent parameterization models -- Variational Autoencoder (VAE, latent space dimensionality: 64), and Latent Bernoulli Autoencoder (LBAE, latent space dimensionality: 256) -- compared against conventional pixel parameterization. The designs were optimized with a BIPOP-CMA-ES optimizer. Note that the shown distributions do not add up to 1 due to the cut-off at 300\%. Best, typical, and worst designs correspond to the best, median, and worst random initializations for each problem.}
  \label{fig:stats2}
\end{figure*}

We found that while optimizing topology without gradients, parameterizing the problem using the latent space of an autoencoder speeds up the process by at least one order of magnitude, closing the gap to gradient-based optimizers reported in the literature. Due to the choice of neural architecture, we demonstrate that the method shows significant robustness and generality, unlike elsewhere in the literature. These findings are demonstrated in Figures \ref{fig:stats1} and \ref{fig:stats2}, which compare our LBAE and VAE architectures against the conventional baseline approach without reparameterization (pixel parameterization). In this example, the designs were optimized with BIPOP-CMA-ES (for complete results, including other optimizers, see the appendix). Performance is quantified as a fraction of problems (the cumulative probability) vs. the relative error of the final solution with respect to the MMA solution -- considered here as the ground truth.

The results are grouped by problem set (in-distribution, and out-of-distribution problems). To investigate the robustness of our method, we also analyze the results considering only subsets of runs -- for each design problem out of 15 randomly initialized optimization runs, we select the best, median, and worst run, to calculate the cumulative probability, as shown in Figures \ref{fig:stats1} and \ref{fig:stats2} in columns 2-4, respectively. In all cases, the LBAE reparameterization significantly outperforms conventional black-box optimization within the prescribed evaluation budget (2000 FEM simulations), regardless of whether we choose all, best, median, or worst runs. In addition, the worst runs of LBAE remain significantly better than the best runs of the conventional pixel parameterization method, proving the overall robustness of this approach. 

As expected, the performance on out-of-distribution problems is worse than on in-distribution problems. At first glance, this can be attributed to the differences between the types of problems seen in the train and test sets. However, worse performance was observed for all the methods, including conventional black-box without reparameterization (done without the use of an autoencoder), which indicates that these out-of-distribution problems are objectively more challenging to optimize. This outcome is particularly encouraging, as it illustrates the difficult task that the proposed method was subjected to.

When comparing LBAE architecture with the VAE, LBAE achieves significantly better performance overall, with the most significant difference visible on the subset of the worst runs, indicating the robustness of our approach. The VAE still achieves a significant margin over the conventional black-box optimization, thus proving the overall benefit of reparameterizing into the latent space, even with a more naive architecture. Nevertheless, the margins between LBAE and VAE indicate that the choice of architecture has a crucial influence on the performance. 

Additionally, we find that the performance in terms of relative errors remains relatively intact whether it is evaluated on grayscale or thresholded designs. This is an important finding, as historically many generative models struggled to produce representations with a significant presence of gray values, which after thresholding leads to disconnected structures (and thus a significantly deteriorated performance) \cite{woldseth2022use}. However, in our case, the designs obtained with LBAE suffer from the gray-value problem to the same extent as the designs obtained with MMA.

Despite a significant advantage over the conventional pixel parameterization, the final objective values when compared to the gradient-based solutions can still be substantial -- some designs show errors on the order of 100\%, i.e. the compliance value is double the one found with MMA. For these problems, the imposed budget of 2000 FEM evaluations turns out to be insufficient to find a good solution. Therefore, unsurprisingly, the designs obtained without gradients remain inferior to those obtained with MMA.

Testing other optimizer configurations (CMA-ES, BIPOP-CMA-ES, with default and customized hyperparameters) led us to the same overall conclusions: latent reparameterization leads to significant improvement over conventional pixel parameterization, and LBAE in particular consistently outperforms the other methods (for details, see the appendix).

\begin{figure*}
    \centering
    \includegraphics[width=0.8\textwidth]{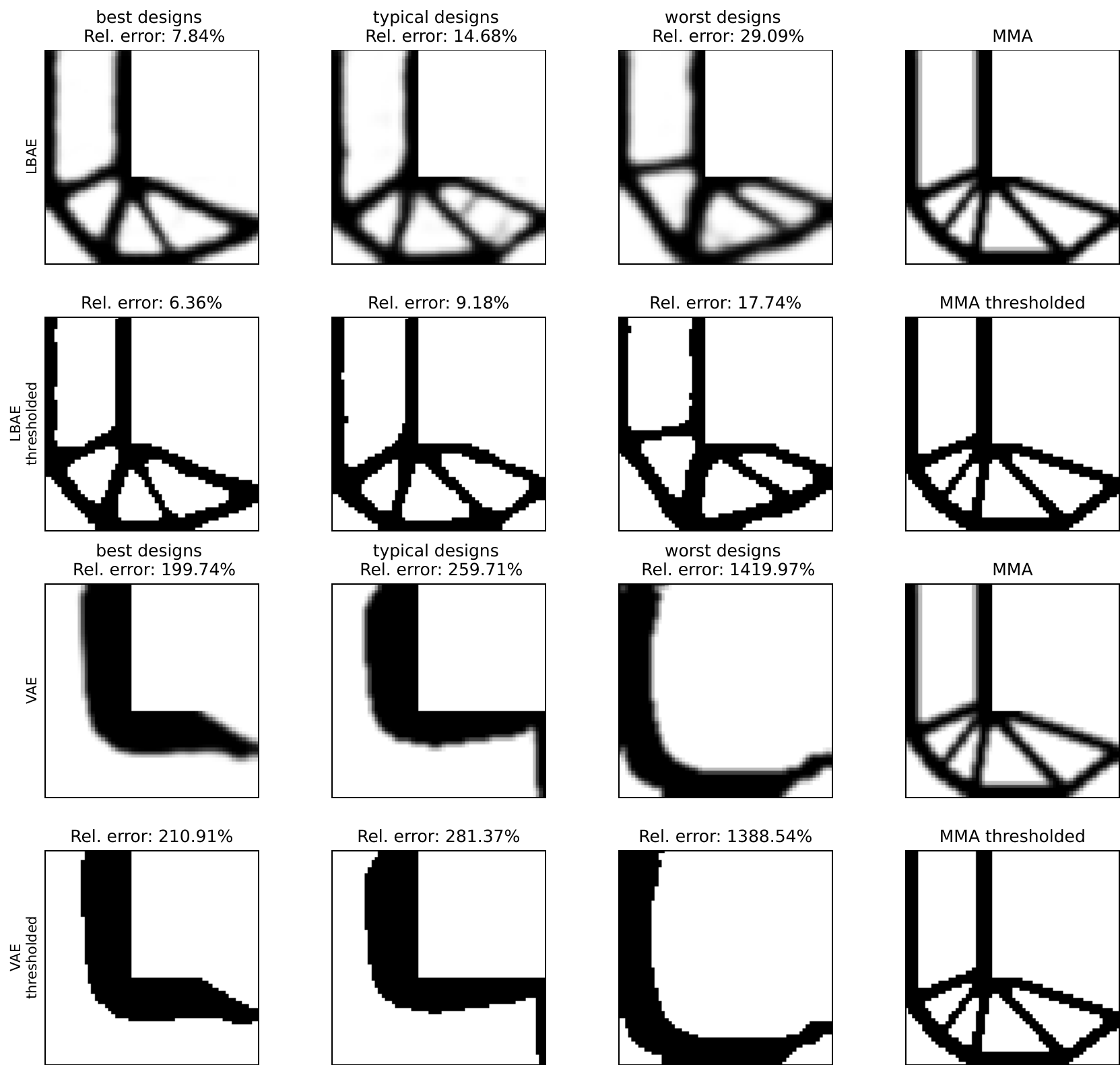}
    \caption{An example of an out-of-distribution problem: classic L-shaped bracket problem, with clamped support along the top edge and a downward point load applied at the right edge. }
    \label{fig:example_solutions}
\end{figure*}

\paragraph{Example designs}
We investigate further the performance of our method by analyzing example designs. Figure \ref{fig:example_solutions} shows an example problem of an L-shaped bracket design from the out-of-distribution set. The example is representative of the most common cases in both problem sets -- the relative errors achieved with LBAE with respect to the MMA solution are relatively low, with the worst solutions being within 10-30\% worse than MMA solutions in terms of the relative error, and the best designs within 5-10\%. The designs are visually quite different from the ground truth and also different from each other -- highlighting the stochastic nature of the gradient-free optimization. Note that for LBAE, the thresholding plays in favor of the model, leading to even lower relative error values compared to the MMA design. In the case of VAE, however, the solutions are significantly oversimplified and lacking detail, which results in significantly worse performance. 

\begin{figure*}
    \centering
    \includegraphics[width=0.8\textwidth]{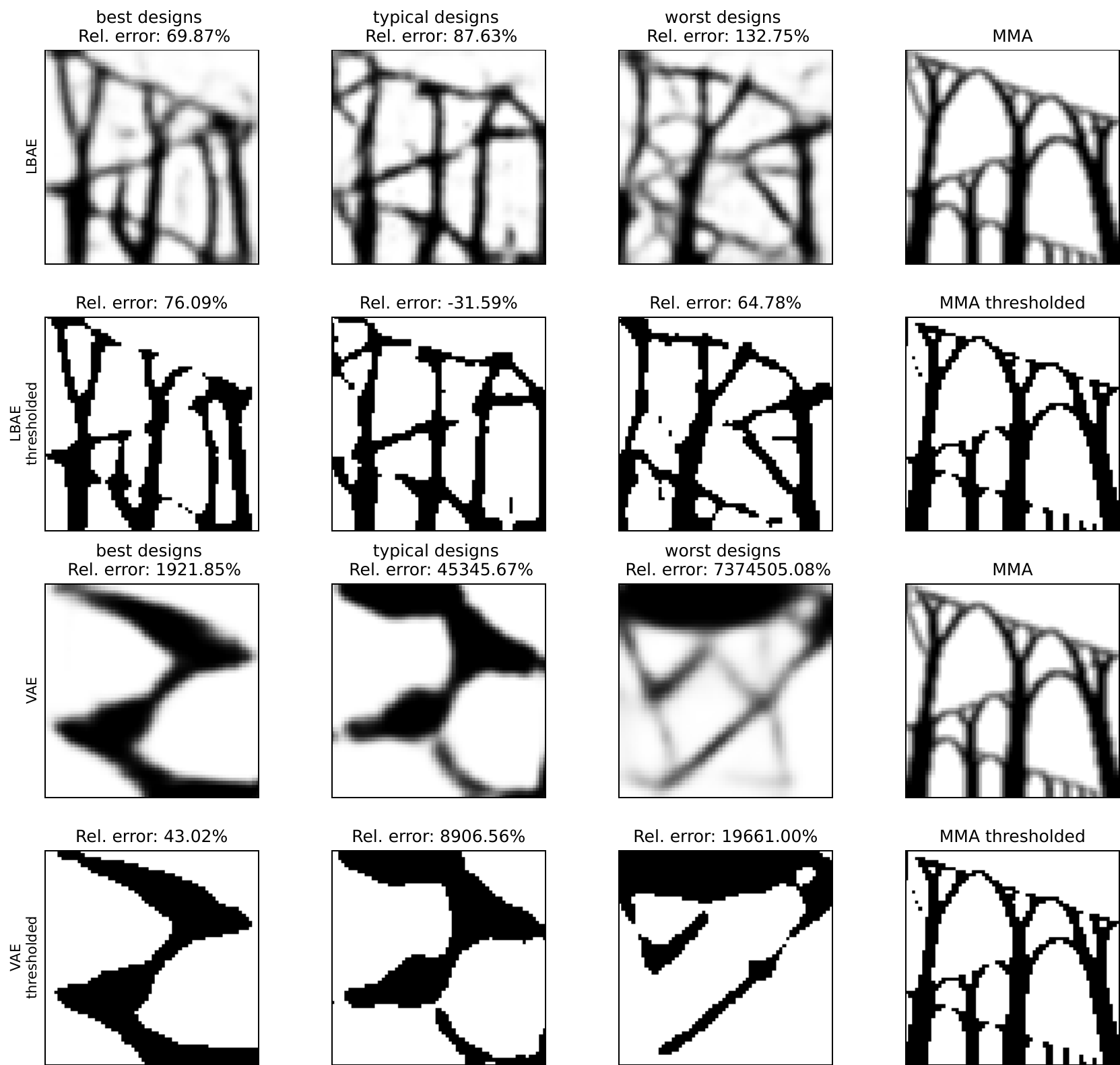}
    \caption{An example of a challenging out-of-distribution problem: a 'staircase' design with Z-shaped distributed load, with clamped support a the bottom edge.}
    \label{fig:example_solutions2}
\end{figure*}

Figure \ref{fig:example_solutions2} shows an example of a more challenging out-of-distribution problem, with a Z-shaped distributed load, requiring a more intricate arrangement of the material. The performance variation is relatively large, with the worst LBAE design yielding 132\% higher compliance compared to the baseline. The designs generated with VAE achieve orders of magnitude worse performance. As in the previous problem (Figure \ref{fig:example_solutions}), the designs obtained with VAE are distinctly oversimplified, indicating the limited capacity of the VAE architecture as compared to LBAE. 

Interestingly, thresholding affects design performance more drastically and inconsistently compared to the first example in Figure \ref{fig:example_solutions}. The reason is the connectivity of the designs and, more specifically, how it is affected by the thresholding step. Noteworthy, some of the designs appear to perform much better after thresholding e.g. typical design in LBAE outperforms the design of MMA. The reason for this is that the design obtained with MMA does suffer to some extent from intermediate 'gray-value' densities present at the elements with prescribed external forces. As a result of thresholding, some of these externally loaded elements become void of material, altering the connectivity and significantly increasing the compliance value of the MMA design. In that case, the placement of the material within the elements subject to external loads in the LBAE design turns out to be more favorable and robust to thresholding, leading to better performance.

Finally, all designs obtained with the conventional black-box optimization approach with pixel parameterization were so far from convergence that we did not include them in the plots. This behavior was observed for all the tested problems in both problem sets, which indicates that the budget of 2000 FEM evaluations is simply insufficient for the conventional black-box optimizer to converge. This observation is in good agreement with the values reported in the literature \cite{sigmund2011usefulness}, stating typical numbers in order of 10,000 - 100,000 FEM iterations until convergence.

\paragraph{Model limitations: heat conduction problem}

In order to explore the limitations of our model in terms of generalization, we deploy our model (LBAE), without retraining, on a fundamentally different problem -- optimization of heat conduction. While optimization for heat conduction can be easily solved with a gradient-based approach, it provides an excellent benchmark for our purpose. Governed by different physics as compared to the structural problems (although it is the same type of partial differential equation), the typical boundary conditions and consequently the solutions are distinctly different. In particular, the design features occurring in thermal problems, are different than those seen in structural compliance optimization (oftentimes slender structures with beam-like elements).

\begin{figure*}[]
    \centering
    \includegraphics[width=0.9\textwidth]{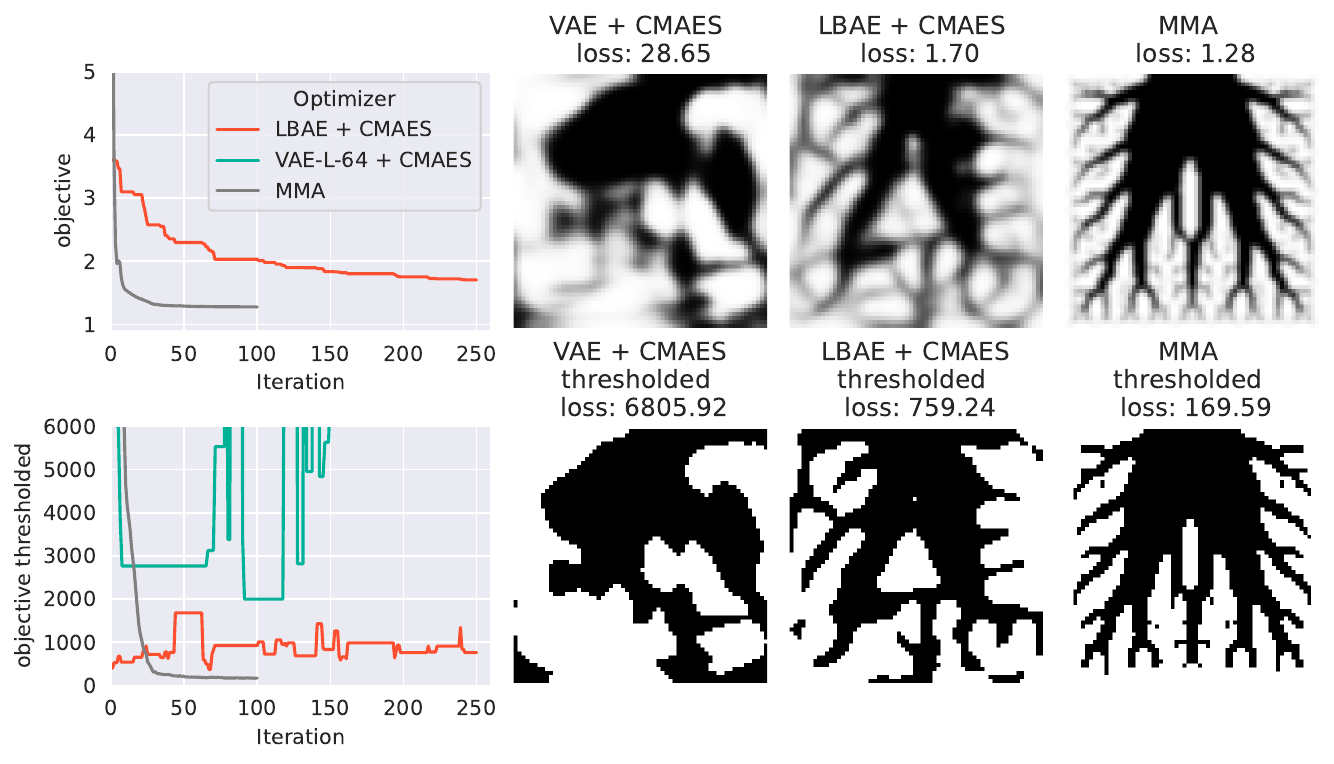}
    \caption{Thermal compliance problem: comparison of latent optimization with CMAES and Latent Bernoulli Autoencoder (LBAE) vs MMA, unprocessed and thresholded solutions. The thresholded objective (bottom plot) was evaluated by thresholding the designs at each iteration, and evaluating the objective. The optimization, however, was carried out using unthresholded designs (top row).}
    \label{fig:thermal1}
\end{figure*}

We set up a simple heat-conduction example following the setup of \cite{liu2014efficient}: there is a uniformly distributed heat source across the square design domain, and a heat sink in the middle of one of the edges, with the length of 0.2 of the edge length. The results obtained with our methods are compared to MMA in Figure \ref{fig:thermal1}. Although our model has never been exposed to thermal problems of any kind, it can produce a solution, with a final objective (1.70) relatively close to that obtained with MMA (1.28). The baseline VAE achieves a significantly worse score (28.65) with a design that lacks finer features, indicating possible limitations regarding VAE's expressivity. 

Additionally, we consider objectives evaluated on thresholded designs, where the densities are rounded to 0 or 1, to eliminate the gray values, while preserving the volume fraction. Thresholding highlights the challenge of this design problem -- the objective is very sensitive to the intermediate density values, making it an interesting benchmark. Due to the uniform heat source, the optimizers attempt to cover as much surface as possible, even at the cost of distributing intermediate densities of the material, leading to gray values. As a result, the value of the objective evaluated on the thresholded design is two orders of magnitude larger, as compared to the objective evaluated on the design that has not been thresholded. When comparing the thresholded designs, latent optimization (LBAE with CMAES) leads to much worse performance, with the objective over four times larger than that obtained with the thresholded MMA design. The thresholded design obtained with VAE yields a much higher (worse) objective than LBAE. 

\begin{figure*}[]
    \centering
    \includegraphics[width=0.9\textwidth]{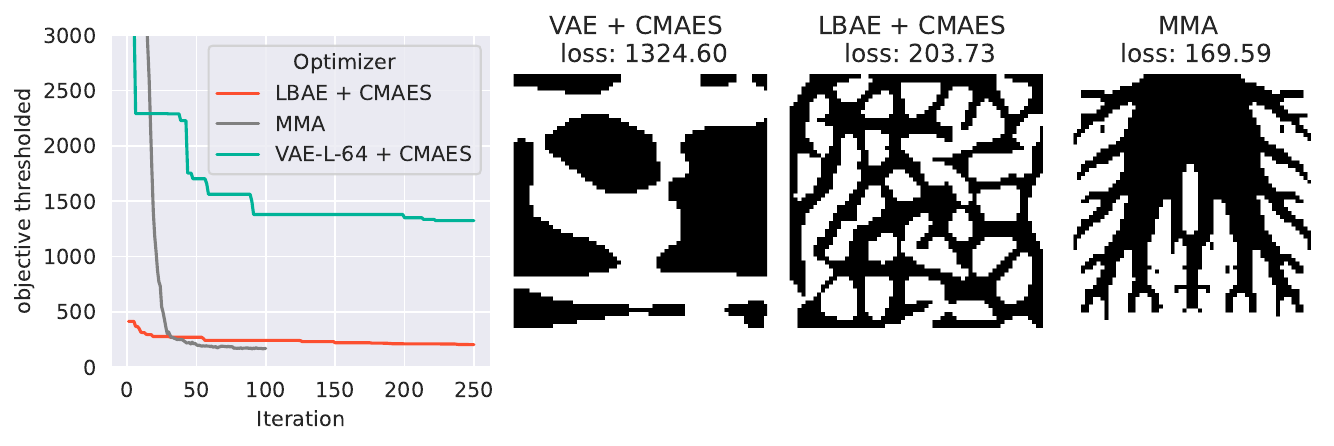}
    \caption{Thermal compliance problem: latent optimization with CMA-ES and Latent Bernoulli Autoencoder (LBAE), optimized for thresholded objective, compared with MMA optimized for un-thresholded objective and, thresholded afterward, in the postprocessing step.}
    \label{fig:thermal2}
\end{figure*}

This problem also offers a good opportunity to demonstrate some advantages of the gradient-free approach -- instead of optimizing for the continuous density design (as required by the gradient-based approach), we can easily change the objective to optimize directly for the thresholded design objective. In other words, instead of optimizing for the objective evaluated with continuous density values, and thresholding in a postprocessing step, we can provide the optimizer with objective values evaluated directly on the thresholded designs. Optimizing for this objective leads to a much better loss value, as shown in Figure \ref{fig:thermal2}, which drops almost four times compared to that of Figure \ref{fig:thermal1}, and is now only 20\% higher than the thresholded MMA design.

\FloatBarrier
\paragraph{Validation: optimizing structure undergoing brittle fracture}

Our main hypothesis for this work is that improving gradient-free topology optimization via neural reparameterization can lead to better results for problems involving non-linear path-dependent physics, with non-smooth objective landscapes, such as brittle fracture. Although topology optimization with consideration of fracture has been addressed in several works \cite{yvonnet2024topology, da2018topology, desai2022topology, russ2019topology, wu2010topology, jia2023controlling}, it remains one of the most challenging problems in structural optimization. The main source of difficulty is the nonlinearity and path-dependence inherent to fracture, which make the objective function highly sensitive to design changes \cite{desai2022topology, russ2019topology}. This poses significant difficulties for gradient-based methods and opens up opportunities for a gradient-free approach. To our knowledge, this is the first time that gradient-free optimization has been applied to optimize structures undergoing brittle fracture. We conjecture that the curse of dimensionality together with the high computational cost of finite element analysis in each iteration has made it unfeasible to use gradient-free approaches in these topology optimization problems. Yet, the proposed gradient-free neural topology optimization strategy might make it feasible, if not even better than \textit{gradient-based} topology optimization strategies for these problems due to the non-smoothness of the objective landscape.

For modeling fracture, we use a phase-field formulation \cite{miehe2010thermodynamically, bourdin2008variational}, as already adopted in the context of topology optimization \cite{yvonnet2024topology}. Compared with alternative approaches such as X-FEM, phase-field facilitates the handling of initiation, propagation, and merging of cracks. It also allows for solving on a fixed mesh which makes it relatively easy to implement and integrate with topology optimization frameworks.

The main principle behind the phase-field method is to model the crack in diffusive form by using an auxiliary scalar field $d \in [0, 1] $, such that the crack surface can be approximated using the following elliptic functional \cite{francfort1998revisiting, miehe2010thermodynamically}:

\begin{align}
    \Gamma_l(d) = \int_{\Omega}\gamma (d, \nabla d) d\Omega
\end{align}

\noindent where $\gamma$ is the crack surface density function, and can be expressed in terms of the phase-field variable $d$ as:   

\begin{align}
        \gamma(d, \nabla d) = \frac{1}{2l}d^2 + \frac{l}{2}\| \nabla d \| ^2
\end{align}

The parameter $l$ is the characteristic length scale that controls the regularization of the problem (crack smearing). With such an approximation, the energy functional over phase-field $d$ and displacement field $\mathbf{u}$ can be formulated as follows:

\begin{align}
    \Pi(\mathbf{u}, d) = \int_{\Omega} \psi(\mathbf{\epsilon}(\mathbf{u}), d)d\Omega + \int_{\Omega} G_c \gamma(d, \nabla d)d\Omega - \int_{\Gamma_N} \mathbf{t}\cdot \mathbf{u}d \Gamma \label{efunc}
\end{align}

where $G_c$ is the Griffith-type energy release rate for a given material, and $\psi$ is the isotropic elastic energy density. The elastic energy density term undergoes an additive decomposition into compressive and tensile components, as shown in Equation (\ref{eq:split0}). 

\begin{align}
     \psi(\mathbf{\epsilon}(\mathbf{u}), d) = \psi^-(\epsilon(\mathbf{u})) + g(d)\psi^+(\mathbf{\epsilon}(\mathbf{u}))\label{eq:split0}
\end{align}

\noindent where the compressive and tensile parts are defined using spectral decomposition of the strain tensor \cite{miehe2010thermodynamically}:

\begin{align}
    \psi^{\pm} = \frac{\lambda}{2} \langle tr[\epsilon(\mathbf{u})]\rangle^2_{\pm} + \mu tr\left[\sum_{i=1}^{dim} \langle \epsilon^i(\mathbf{u})\rangle_{\pm} \mathbf{n^i}\otimes \mathbf{n^i}\right]^2 \label{split}
\end{align}

Here $\langle\rangle_{\pm}$ denote the Macaulay brackets, defined as $\langle x\rangle_{+} = \max(0, x)$ and $\langle x\rangle_{-} = \min(0, x)$. The terms $\mathbf{n^i}$ and $\epsilon^i(\mathbf{u})$ are respectively the eigenvectors and eigenvalues of the strain tensor, and $g(d)$ is the degradation function, which takes the form: $g(d) = (1-d)^2$. The purpose of spectral decomposition into tensile and compressive parts is to ensure that the crack growth can be induced only in tension -- as can be seen in Equation (\ref{split}), the degradation function term affects only the tensile term. 

The strong form can be obtained by minimizing the energy stated in Equation (\ref{efunc}), such that $\delta \Pi = 0$. This results in the following coupled PDE system: 

\begin{align}
    \nabla \cdot \mathbf{\sigma}(\mathbf{u}, d) = \mathbf{0} && \text{in } \Omega\label{eq6}\\
    \frac{G_c}{l}\left(d - l^2 \Delta d\right) = 2(1 - d) \mathcal{H} && \text{in } \Omega\label{eq7}\\
    \mathbf{n} \cdot  \mathbf{\sigma}(\mathbf{u}, d) = \mathbf{t} && \text{on } \partial \Omega^t\label{eq8}\\
    \mathbf{u} = \mathbf{u^D} && \text{on }  \partial\Omega^u \label{eq8}\\
    \mathbf{n} \cdot \nabla d = 0 && \text{on }  \partial \Omega \label{eq9}\\
\end{align}

Equation (\ref{eq6}) represents the quasi-static force balance, while Equation (\ref{eq7}) governs the phase field evolution. Here $\mathcal{H}$ is the history variable, defined as $\mathcal{H}(\mathbf{x}, t) = \max_{\tau \in [0, t]}\left[\psi_e^+(\mathbf{x}, t)\right]$, which enforces irreversibility of the damage or, in other words, prevents crack healing. In Equation (\ref{eq8}) the term $\mathbf{n}$ corresponds to the surface normal vector, $\mathbf{t}$ represents the traction applied on the boundary $\partial \Omega^{t}$. The term $\mathbf{u}^D$ corresponds to the Dirichlet boundary conditions prescribed on the boundary $\partial \Omega ^{u}$. 

This strong form gives rise to a weak form that is discretized using the finite element method (for details, see e.g. \cite{bourdin2008variational, russ2019topology, desai2022topology}) and solved numerically. Due to the path dependence, the loading is applied in increments. The solutions for $\mathbf{u}$ and $d$ are found by solving the coupled systems of the FEM equations using the standard iterative staggered scheme, which is the most common approach \cite{russ2019topology, da2018topology, desai2022topology}. Furthermore, the displacement equations, which are non-linear due to material degradation, are solved using the iterative Newton method. More details on the parameters used in this fracture simulation can be found in the appendix.

The optimization problem is formulated as maximizing the total external work, with maximum volume constraint, as shown in Equation (\ref{eq:optimization}). The formulation is akin to that posed in \cite{desai2022topology, da2018topology}.

\begin{align}
 \left. \begin{matrix}
    \text{max:} & W_e = \int_{\Omega}{\epsilon(\mathbf{u},d) : \sigma(\mathbf{u}, d)d\Omega} \\
    \text{s.t.:} & \int_{\Omega}{\rho}dV = V_0\\
    & 0 < \rho < 1 \\
    \end{matrix}
    \right \}
    \label{eq:optimization}
\end{align}

\begin{figure}[h]
    \centering
    \includegraphics[width=0.65\linewidth]{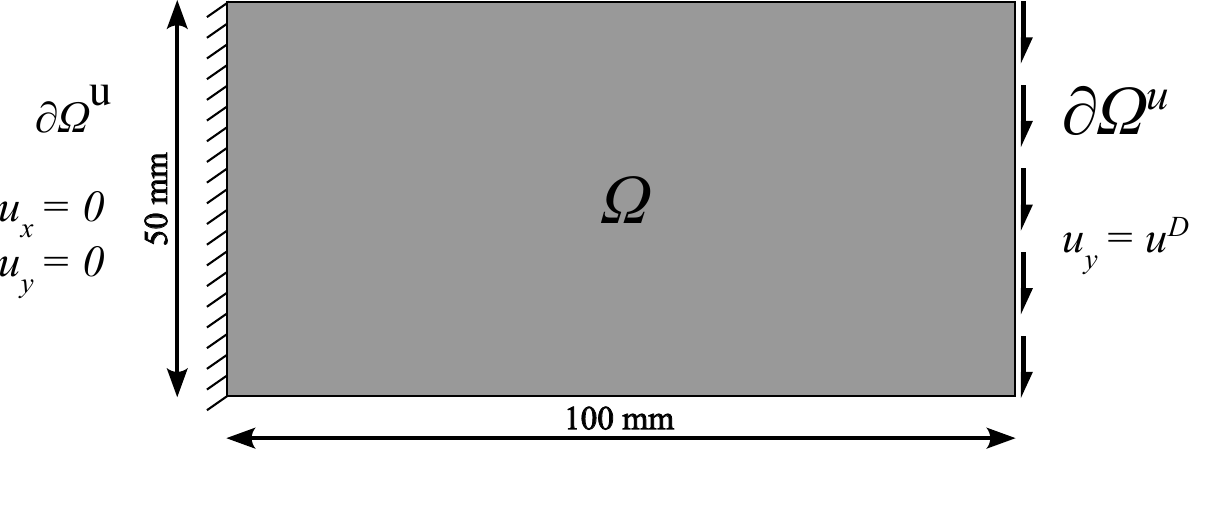}
    \caption{Illustration of the boundary conditions of the cantilever beam problem. The left edge is clamped, while at the right edge, the load acting downwards is applied in terms of the prescribed displacement $u^D$. The displacement is applied in several increments until it reaches a maximum of 0.5 mm. }
    \label{fig:bcs}
\end{figure}

The parameterization of material properties is based on the SIMP method, following the approach of \cite{russ2019topology}. While using SIMP comes with certain limitations \cite{russ2019topology}, such as non-physical `gray-values', it remains a common practice in the topology optimization community \cite{jia2023controlling, russ2019topology}. To regularize the solution and prevent checkerboard patterns, we apply a standard cone filter \cite{88line}. The boundary conditions for our benchmark problem correspond to a simple cantilever beam, as shown in Figure \ref{fig:bcs}, that is based on examples of \cite{desai2022topology} and \cite{russ2019topology}.

We solve this optimization problem using our proposed approach of gradient-free neural topology optimization and compare it with a traditional solution method, i.e. using a gradient-based MMA algorithm. Importantly, \textit{we use our LBAE model without retraining}, i.e. it has been trained only once on the dataset comprised of designs optimized for compliance. Furthermore, to match the shape and size of the discretized domain of our benchmark (64x128 elements) we resize the output of the LBAE model using standard bi-linear interpolation. We set the CMA-ES population size to 16, which means that in each iteration 16 candidate solutions are evaluated (in parallel). For the baseline gradient-based solution we use sensitivities obtained with automatic differentiation based on JAX \cite{xue2023jax, jax2018github}.

Both gradient-based and gradient-free approaches were evaluated using two different initialization strategies. In the first scenario, we use the default initialization, i.e. for MMA the initial design was set to a uniform density field satisfying the volume fraction constraint, while the initial CMA-ES distribution was initialized using the default $\mathbf{0}$ mean. In the second strategy, the optimizers were initialized using a design optimized for minimum compliance using a linear elasticity model. In the case of CMA-ES, this initial design was encoded into the latent vector $\mathbf{z}$ using the encoder, and the latent vector was used to set the mean of the initial distribution. Additionally, for the gradient-free optimizer, we tested different initial values for $\sigma_0$, which controls the spread in the initial distribution.

\begin{figure}
    \centering
    \includegraphics[width=1.0\linewidth]{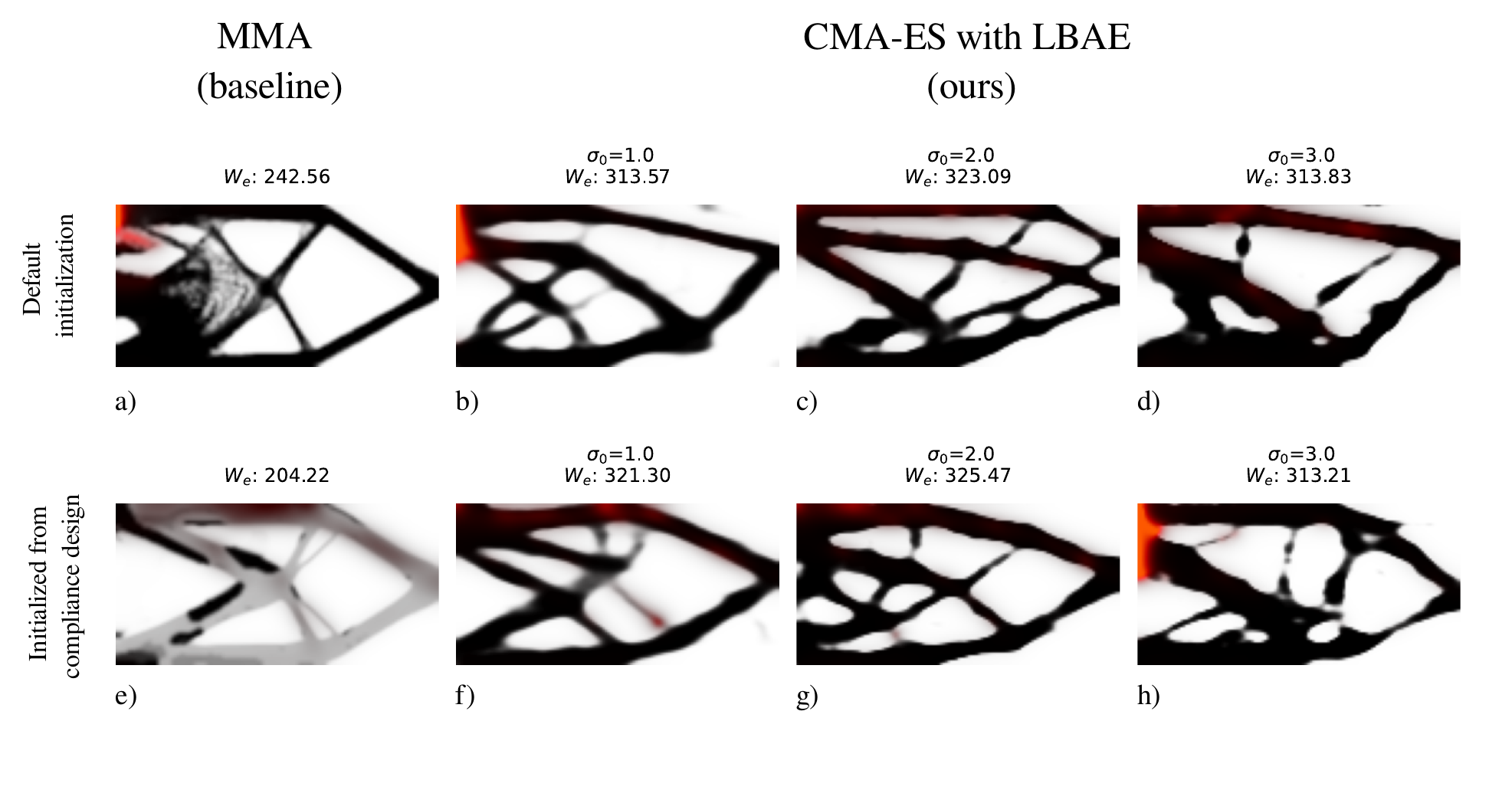}
    \caption{Comparison of final designs obtained with our proposed gradient-free approach - gradient-free CMA-ES with LBAE reparameterization compared against the baseline designs obtained with gradient-based MMA optimizer. The results were obtained for two different initialization strategies: default (uniform for MMA and $\mathbf{0}$ mean for CMA-ES) as well as starting from the design optimized for compliance governed by linear elasticity (in that case, the mean of CMA-ES distribution is initialized from the compliance design encoded into a latent vector). Furthermore, CMA-ES was tested for 3 different settings of the initial variance parameter $\sigma_0$ for the initial population distribution. }
    \label{fig:frac-designs}
\end{figure}

The results are summarized in Figure \ref{fig:frac-designs}, which presents the final design and the corresponding objective values. As shown in the figure, gradient-free optimization outperforms the gradient-based approach by a significant margin (around 30\%) in all tested configurations. Investigating the designs obtained with CMA-ES, it can be seen that in all cases the optimizer managed to identify a better design strategy, reinforcing the upper left corner, where the tensile loads are maximum, and in this way postponing fracture. Following this strategy, the designs are characterized by an asymmetric layout -- a typical feature observed in designs optimized with consideration of fracture \cite{desai2022topology, russ2019topology}.

On the contrary, the gradient-based optimizer struggled to identify this strategy, due to getting stuck in local optima. In the case of initialization from the compliance design (Figure \ref{fig:frac-designs}e), the optimizer gets stuck in a local minimum which exploits the SIMP parameterization. In an attempt to reduce the damage due to fracture, the optimizer lowers the density values throughout the whole domain, leading to lower stiffness and lower stresses, in such a way attempting to prevent damage. This leads to the "gray values" design with a relatively low stiffness (and thus lower external work needed to reach the prescribed displacement), which results in mediocre overall performance. However, while attempting to increase the density values, damage is induced in the design, resulting in a `barrier' in the loss landscape, which the optimizer is unable to cross.

\begin{figure*}
    \centering
    \includegraphics[width=0.9\linewidth]{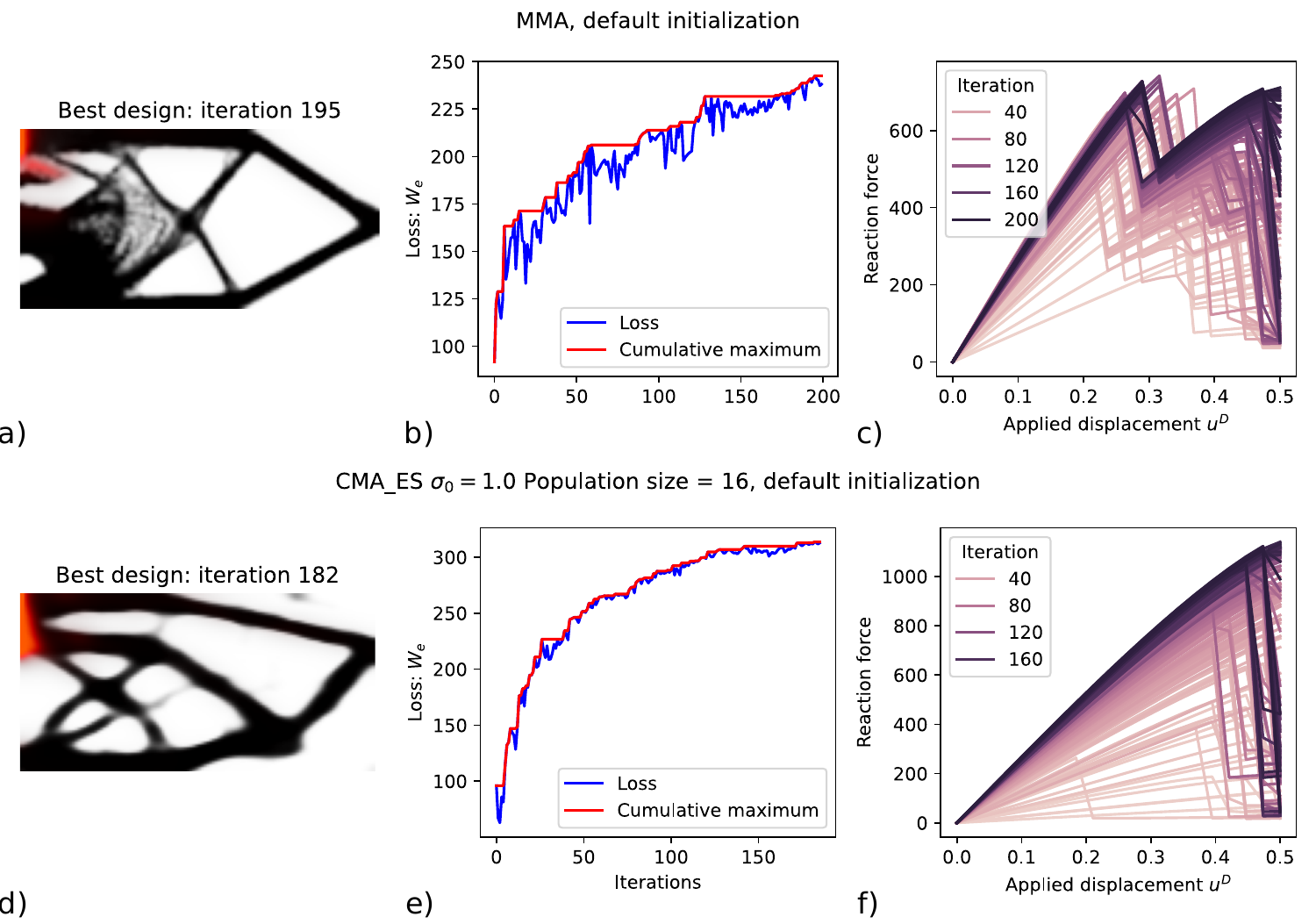}
    \caption{Comparison of optimization with fracture using gradient-based MMA vs gradient-free CMA-ES with LBAE reparameterization. Pictures a) and d) represent the best designs with the phase-field damage marked in red. For CMA-ES in each iteration, the whole population is evaluated, i.e. 16 designs are simulated in parallel. }
    \label{fig:frac-combined}
\end{figure*}

In the second case (Figure \ref{fig:frac-designs}a) the evidence of the optimizer getting stuck in the local minimum is more subtle, yet it can be found by investigating the force-displacement curves shown in Figure \ref{fig:frac-combined}. Unlike the design obtained with CMA-ES (Figure \ref{fig:frac-combined}d), which is reinforced in the upper left corner, the structure obtained with MMA (Figure \ref{fig:frac-combined}a) is allowed to fracture in the upper corner relatively quickly, and instead, the optimizer attempts to reinforce the remaining intact part (lower left corner) - or, in other words, improve the post-fracture response of the design. This manifests itself in the force-displacement curve as a drop in the reaction force at the displacement of $u^D = 0.3$ mm. However, this approach leads to a suboptimal design with around 30\% lower strain energy absorption, as opposed to the CMA-ES solution. As can be seen from the history of the force-displacement curves, this characteristic saw-tooth response pattern is present starting in the early iterations. It appears that the MMA optimizer is not able to change the response of the design, and gets stuck in a suboptimal reinforcement strategy. 

\section{Discussion}

\paragraph{Model limitations}

We observe that the designs generated by LBAE oftentimes tend to be simpler, than those obtained with MMA (see Figure \ref{fig:example_solutions}). While in certain scenarios this can be a desired feature (e.g. for manufacturing), it appears that in this case, the simplicity originates from some deficiencies of the generator model, which struggles with producing finer features. 

\begin{figure*}[h]
    \centering
    \includegraphics[width=\textwidth]{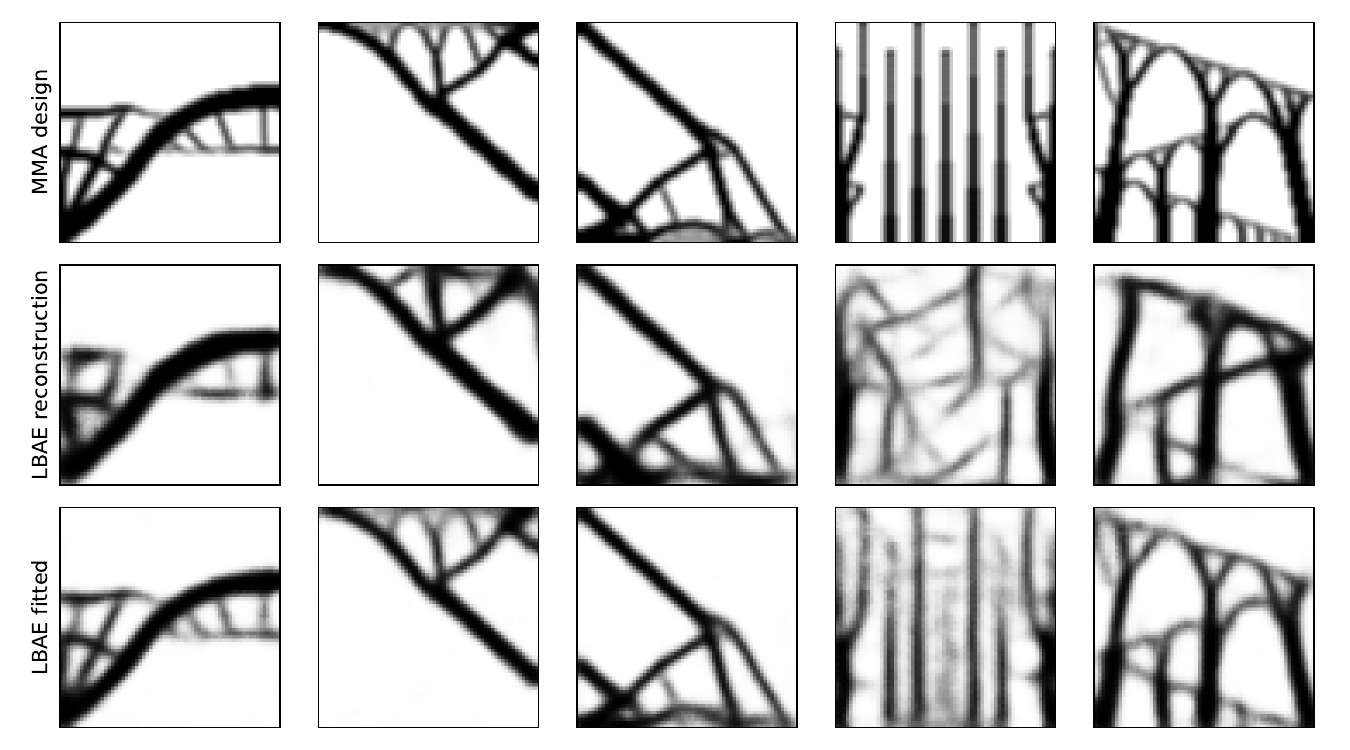}
    \caption{Examples of baseline designs obtained with MMA, and their reconstructions obtained with LBAE. Note that LBAE still exhibits some blurriness when reconstructing finer features. }
    \label{fig:expressivity_check}
\end{figure*}

To verify these limitations, we performed a simple study on the expressivity of the model. We consider several sample designs (obtained with MMA) and pass them through the LBAE, to investigate whether the model has enough expressivity for reconstruction. As shown in Figure \ref{fig:expressivity_check}, the designs reconstructed from the latent encoding provided by the encoder are imperfect and lack the finer features. Next, we take the latent encoding of each design and optimize it (fitting MSE loss to the MMA design, using ADAM optimizer), to verify whether the model has enough expressive capacity. The results of the decoded designs with fitted latent vectors are shown in the bottom row. These designs incorporate finer features, providing more accurate and detailed representations of the MMA target designs. Nevertheless, the blurriness of these finer features is not eliminated, indicating the expressivity limits of the decoder.  

We speculate that these effects could be partially attributed to the convolutional layers -- known to exhibit higher impedance to finer features \cite{ulyanov2018deep, rahaman2019spectral} and the spectral bias of neural networks \cite{rahaman2019spectral}. In other words, convolutional neural nets struggle more with fitting fine features than coarse ones. This deficiency, however, could be potentially addressed by extending the training, or further adjustments in the architecture. As shown in the examples presented in Figure \ref{fig:example_solutions}, a lack of fine features can have a detrimental influence on the performance of the designs and thus the overall optimization capabilities of the proposed method. Addressing those limitations might lead to significant improvements in the model's performance.  

Finally, the approach using variational autoencoder architectures based on convolutional neural networks is by design constrained to uniform meshes, and restricted in terms of resolution. Furthermore, we have not tried to solve 3D problems, and we do not know how the approach would scale in those scenarios. 

\paragraph{Optimization limitations}

The second component of the performance of our method comes from the optimization procedure and its limitations. The major point of attention in this regard is the setting of the constraints. Specifically, the volume constraint was imposed following the method of \cite{hoyer2019neural}, by applying a constrained sigmoid transformation on the output of the convolutional layers of the generator. However, this way of enforcing constraints could potentially hinder latent optimization. During training, the VAE was exposed to samples with different volume fractions, resulting in not only different shapes but also different topologies. Therefore, imposing the constraint directly in latent space by restricting to a manifold corresponding to a particular volume fraction could be more beneficial. Alternatively, the volume fraction could be an input to the model. In this way, the model would be trained to output designs that satisfy (approximately) the volume constraint. Finding a different way of enforcing the constraint could potentially further improve the performance of the proposed latent optimization method. 

\begin{figure*}
    \centering
    \includegraphics[width=\textwidth]{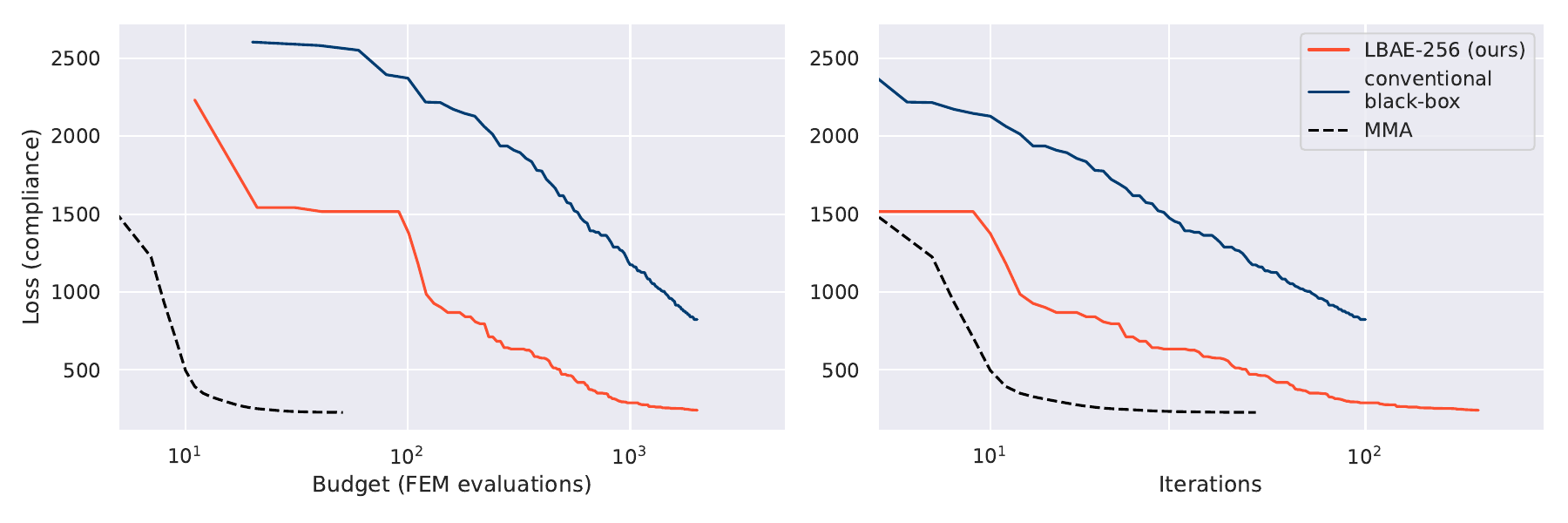}
    \caption{Comparison of loss curves for an example out of distribution problem. Given that for gradient-free optimization the simulations can be executed in parallel for each population, comparing performance based on iteration is more representative of the actual time (and not the computational cost, as with comparison by FEM evaluations). From this perspective, the gap between gradient-based and gradient-free optimizers closes even further. }
    \label{fig:loss3}
\end{figure*}

\paragraph{Benchmarking}
In our benchmarks, we compare different methods based on the budget of objective function evaluations (number of simulations) and optimizer iterations. In a single iteration, gradient-free optimizers usually evaluate a whole population of trial solutions, i.e. a simulation is run for each trial solution. However, these simulations are run in parallel, which means that the iteration time remains comparable to the compute time of a single simulation (assuming that the FEM simulation comprises the majority of the computational cost in the optimization loop). In other words, the budget (number of simulations) is more representative of the computational resources (CPU time), while the number of iterations is more representative of the absolute compute time. An example comparison between different methods using both metrics is shown in Figure \ref{fig:loss3}. Since the population size in our set-up is on the order of 10, the difference between budget and iterations for gradient-free approaches is approximately 1 order of magnitude. In the gradient-based approach, there is a single simulation per iteration (although in practice, in general, the cost of computing sensitivities can be comparable to the cost of the simulation). 

Although our approach significantly closes the gap between gradient-free optimization without parameterization and gradient-based methods, as shown in Figure \ref{fig:loss3}, it still remains relatively expensive in terms of computational resources for smooth and differentiable problems. We do not expect this approach to be competitive on problems that can be solved efficiently with gradient-based methods. However, with our validation example, we show that our method enables gradient-free optimization with reasonable computational resources and even achieves better objective values for problems where gradient-based algorithms struggle.

Finally, our validation example is relatively limited. It can be argued that the issues of the gradient-based optimizer can be overcome with a number of techniques such as e.g. continuation schemes, hyperparameter tuning, or alternative formulations of the optimization problems \cite{russ2019topology, desai2022topology}. Alternatively, these challenges could be tackled with an emerging approach that involves the use of Physics-Informed Gaussian Processes, which optimizes the solution of the governing PDEs simultaneously with the design variables, thereby eliminating the need for continuation schemes \cite{yousef2024}. While with these techniques better performance may be reached, we also want to highlight the robustness of our method to the choice of hyperparameters and initialization -- an important feature from the practitioner's perspective. Nevertheless, future work is encouraged to further assess the merits of our method.

\section{Conclusion}

In this study, we presented a framework for optimizing topology designs without gradients in a latent space of a Latent-Bernoulli Autoencoder (LBAE). We found that the proposed method significantly improves the scalability of gradient-free optimization -- addressing the major shortcoming of this approach. 

Our extensive benchmarks indicate that reparameterizing the optimization problem into a lower-dimensional latent space of LBAE leads to dramatic gains in performance, with at least one order of magnitude faster convergence as opposed to the conventional black-box optimization without reparameterization. We demonstrated the robustness of our method, by showing that the performance gains are significant even in the worst-case scenario i.e. considering only the worst runs out of all random initialization. 

We applied our method without retraining to optimize a structure undergoing brittle fracture and demonstrated that it is capable of outperforming the gradient-based approach on problems characterized by discontinuous loss landscapes, where gradient-based optimizers tend to struggle. Specifically, our proposed method outperformed the traditional gradient-based optimizer (MMA) by a significant margin, consistently delivering an improvement of the order of 30\% on all tested configurations with different initialization strategies and hyperparameter settings. 

Nevertheless, our implementation still suffers from several limitations. We decided to impose hard constraints (volume fraction), which potentially hinder the optimization performance. We speculate that imposing constraints directly in the latent space might potentially lead to better optimization. The results indicate a bias towards certain feature length scales in the designs. Generating features over a wider range of length scales is desired to improve the expressivity and variety of the solutions. Our current architecture is limited to uniform grid meshes and a restricted resolution. Convolutional layers may not scale well to 3D applications. Therefore, exploring different architectures is expected to lead to even more significant performance gains, minimize the dimension of the latent space, further improve the scalability, and extend the applicability of latent space optimization. Finally, we limited our study to density-based optimization and volume fraction constraint. Exploring other parameterization approaches, such as level-set method, and other types of constraints, remains a subject for future work. 

Still, we are surprised by the large performance improvements we found in our work relative to a conventional gradient-free approach to topology optimization, and we find it particularly encouraging to show for the first time the generalization ability on problems, where the traditional gradient-based optimizers reach their limitations.

\section*{Acknowledgments}
The authors would like to acknowledge that this effort was undertaken in part with the support from the Department of the Navy, Office of Naval Research, award number N00014-23-1-2688. Part of this research was conducted using computational
resources and services at the Center for Computation and Visualization, Brown University.

\bibliographystyle{unsrt}  
\bibliography{references}  

\begin{thebibliography}{10}

\bibitem{sigmund2013topology}
Ole Sigmund and Kurt Maute.
\newblock Topology optimization approaches: A comparative review.
\newblock {\em Structural and multidisciplinary optimization},
  48(6):1031--1055, 2013.

\bibitem{munk2019benefits}
David~J Munk, Douglass~J Auld, Grant~P Steven, and Gareth~A Vio.
\newblock On the benefits of applying topology optimization to structural
  design of aircraft components.
\newblock {\em Structural and Multidisciplinary Optimization}, 60:1245--1266,
  2019.

\bibitem{zhu2016topology}
Ji-Hong Zhu, Wei-Hong Zhang, and Liang Xia.
\newblock Topology optimization in aircraft and aerospace structures design.
\newblock {\em Archives of computational methods in engineering}, 23:595--622,
  2016.

\bibitem{papadrakakis1998structural}
Manolis Papadrakakis, Nikos~D Lagaros, and Yiannis Tsompanakis.
\newblock Structural optimization using evolution strategies and neural
  networks.
\newblock {\em Computer methods in applied mechanics and engineering},
  156(1-4):309--333, 1998.

\bibitem{xie1993simple}
Yi~Min Xie and Grant~P Steven.
\newblock A simple evolutionary procedure for structural optimization.
\newblock {\em Computers \& structures}, 49(5):885--896, 1993.

\bibitem{woldseth2022use}
Rebekka~V Woldseth, Niels Aage, J~Andreas B{\ae}rentzen, and Ole Sigmund.
\newblock On the use of artificial neural networks in topology optimisation.
\newblock {\em Structural and Multidisciplinary Optimization}, 65(10):294,
  2022.

\bibitem{wu2021topology}
Jun Wu, Ole Sigmund, and Jeroen~P Groen.
\newblock Topology optimization of multi-scale structures: a review.
\newblock {\em Structural and Multidisciplinary Optimization}, 63:1455--1480,
  2021.

\bibitem{sigmund200199}
Ole Sigmund.
\newblock A 99 line topology optimization code written in matlab.
\newblock {\em Structural and multidisciplinary optimization}, 21:120--127,
  2001.

\bibitem{desai2022topology}
Jeet Desai, Gr{\'e}goire Allaire, and Fran{\c{c}}ois Jouve.
\newblock Topology optimization of structures undergoing brittle fracture.
\newblock {\em Journal of Computational Physics}, 458:111048, 2022.

\bibitem{da2018topology}
Daicong Da, Julien Yvonnet, Liang Xia, and Guangyao Li.
\newblock Topology optimization of particle-matrix composites for optimal
  fracture resistance taking into account interfacial damage.
\newblock {\em International Journal for Numerical Methods in Engineering},
  115(5):604--626, 2018.

\bibitem{schwarz2001topology}
Stefan Schwarz, Kurt Maute, and Ekkehard Ramm.
\newblock Topology and shape optimization for elastoplastic structural
  response.
\newblock {\em Computer methods in applied mechanics and engineering},
  190(15-17):2135--2155, 2001.

\bibitem{guirguis2019evolutionary}
David Guirguis, Nikola Aulig, Renato Picelli, Bo~Zhu, Yuqing Zhou, William
  Vicente, Francesco Iorio, Markus Olhofer, Wojciech Matusik, Carlos
  Artemio~Coello Coello, et~al.
\newblock Evolutionary black-box topology optimization: Challenges and
  promises.
\newblock {\em IEEE Transactions on Evolutionary Computation}, 24(4):613--633,
  2019.

\bibitem{sigmund2011usefulness}
Ole Sigmund.
\newblock On the usefulness of non-gradient approaches in topology
  optimization.
\newblock {\em Structural and Multidisciplinary Optimization}, 43:589--596,
  2011.

\bibitem{hansen2010comparing}
Nikolaus Hansen, Anne Auger, Raymond Ros, Steffen Finck, and Petr
  Po{\v{s}}{\'\i}k.
\newblock Comparing results of 31 algorithms from the black-box optimization
  benchmarking bbob-2009.
\newblock In {\em Proceedings of the 12th annual conference companion on
  Genetic and evolutionary computation}, pages 1689--1696, 2010.

\bibitem{russ2019topology}
Jonathan~B Russ and Haim Waisman.
\newblock Topology optimization for brittle fracture resistance.
\newblock {\em Computer Methods in Applied Mechanics and Engineering},
  347:238--263, 2019.

\bibitem{huang2023topology}
Hua-Ming Huang, Elena Raponi, Fabian Duddeck, Stefan Menzel, and Mariusz Bujny.
\newblock Topology optimization of periodic structures for crash and static
  load cases using the evolutionary level set method.
\newblock {\em Optimization and Engineering}, pages 1--34, 2023.

\bibitem{guo2018indirect}
Tinghao Guo, Danny~J Lohan, Ruijin Cang, Max~Yi Ren, and James~T Allison.
\newblock An indirect design representation for topology optimization using
  variational autoencoder and style transfer.
\newblock In {\em 2018 AIAA/ASCE/AHS/ASC Structures, Structural Dynamics, and
  Materials Conference}, page 0804, 2018.

\bibitem{88line}
Erik Andreassen, Anders Clausen, Mattias Schevenels, Boyan~S Lazarov, and Ole
  Sigmund.
\newblock Efficient topology optimization in matlab using 88 lines of code.
\newblock {\em Structural and Multidisciplinary Optimization}, 43:1--16, 2011.

\bibitem{hoyer2019neural}
Stephan Hoyer, Jascha Sohl-Dickstein, and Sam Greydanus.
\newblock Neural reparameterization improves structural optimization.
\newblock {\em arXiv preprint arXiv:1909.04240}, 2019.

\bibitem{zhang2021tonr}
Zeyu Zhang, Yu~Li, Weien Zhou, Xiaoqian Chen, Wen Yao, and Yong Zhao.
\newblock Tonr: An exploration for a novel way combining neural network with
  topology optimization.
\newblock {\em Computer Methods in Applied Mechanics and Engineering},
  386:114083, 2021.

\bibitem{jameson2007adjoint}
Sriram Jameson and Antony Jameson.
\newblock Adjoint formulations for topology, shape and discrete optimization.
\newblock In {\em 45th AIAA Aerospace Sciences Meeting and Exhibit}, page~55,
  2007.

\bibitem{giraldo2020unified}
Oliver Giraldo-Londo{\~n}o and Glaucio~H Paulino.
\newblock A unified approach for topology optimization with local stress
  constraints considering various failure criteria: von mises, drucker--prager,
  tresca, mohr--coulomb, bresler--pister and willam--warnke.
\newblock {\em Proceedings of the Royal Society A}, 476(2238):20190861, 2020.

\bibitem{zhao2019material}
Tuo Zhao, Adeildo~S Ramos~Jr, and Glaucio~H Paulino.
\newblock Material nonlinear topology optimization considering the von mises
  criterion through an asymptotic approach: Max strain energy and max load
  factor formulations.
\newblock {\em International Journal for Numerical Methods in Engineering},
  118(13):804--828, 2019.

\bibitem{zhang2020topology}
Zeyu Zhang, Yong Zhao, Bingxiao Du, Xiaoqian Chen, and Wen Yao.
\newblock Topology optimization of hyperelastic structures using a modified
  evolutionary topology optimization method.
\newblock {\em Structural and Multidisciplinary Optimization}, 62:3071--3088,
  2020.

\bibitem{balamurugan2008performance}
R~Balamurugan, CV~Ramakrishnan, and Nidur Singh.
\newblock Performance evaluation of a two stage adaptive genetic algorithm
  (tsaga) in structural topology optimization.
\newblock {\em Applied Soft Computing}, 8(4):1607--1624, 2008.

\bibitem{balamurugan2011two}
R~Balamurugan, CV~Ramakrishnan, and N~Swaminathan.
\newblock A two phase approach based on skeleton convergence and geometric
  variables for topology optimization using genetic algorithm.
\newblock {\em Structural and Multidisciplinary Optimization}, 43:381--404,
  2011.

\bibitem{ramamoorthy2023multi}
Vivek~T Ramamoorthy, Ender {\"O}zcan, Andrew~J Parkes, Luc Jaouen, and
  Fran{\c{c}}ois-Xavier B{\'e}cot.
\newblock Multi-objective topology optimisation for acoustic porous materials
  using gradient-based, gradient-free, and hybrid strategies.
\newblock {\em The Journal of the Acoustical Society of America},
  153(5):2945--2945, 2023.

\bibitem{luh2004multi}
Guan-Chun Luh and Chung-Huei Chueh.
\newblock Multi-modal topological optimization of structure using immune
  algorithm.
\newblock {\em Computer Methods in Applied Mechanics and Engineering},
  193(36-38):4035--4055, 2004.

\bibitem{kaveh2008structural}
Ali Kaveh, Behrooz Hassani, S~Shojaee, and Shahariar~Mehdi Tavakkoli.
\newblock Structural topology optimization using ant colony methodology.
\newblock {\em Engineering Structures}, 30(9):2559--2565, 2008.

\bibitem{luh2009structural}
Guan-Chun Luh and Chun-Yi Lin.
\newblock Structural topology optimization using ant colony optimization
  algorithm.
\newblock {\em Applied Soft Computing}, 9(4):1343--1353, 2009.

\bibitem{luh2011binary}
Guan-Chun Luh, Chun-Yi Lin, and Yu-Shu Lin.
\newblock A binary particle swarm optimization for continuum structural
  topology optimization.
\newblock {\em Applied Soft Computing}, 11(2):2833--2844, 2011.

\bibitem{plevris2011hybrid}
Vagelis Plevris and Manolis Papadrakakis.
\newblock A hybrid particle swarm—gradient algorithm for global structural
  optimization.
\newblock {\em Computer-Aided Civil and Infrastructure Engineering},
  26(1):48--68, 2011.

\bibitem{shim1997generating}
Patrick~Y Shim and Souran Manoochehri.
\newblock Generating optimal configurations in structural design using
  simulated annealing.
\newblock {\em International journal for numerical methods in engineering},
  40(6):1053--1069, 1997.

\bibitem{lee2004new}
Kang~Seok Lee and Zong~Woo Geem.
\newblock A new structural optimization method based on the harmony search
  algorithm.
\newblock {\em Computers \& structures}, 82(9-10):781--798, 2004.

\bibitem{wu2010topology}
Chun-Yin Wu and Ko-Ying Tseng.
\newblock Topology optimization of structures using modified binary
  differential evolution.
\newblock {\em Structural and Multidisciplinary Optimization}, 42:939--953,
  2010.

\bibitem{lu2018structured}
Xiaoyu Lu, Javier Gonzalez, Zhenwen Dai, and Neil~D Lawrence.
\newblock Structured variationally auto-encoded optimization.
\newblock In {\em International conference on machine learning}, pages
  3267--3275. PMLR, 2018.

\bibitem{notin2021improving}
Pascal Notin, Jos{\'e}~Miguel Hern{\'a}ndez-Lobato, and Yarin Gal.
\newblock Improving black-box optimization in vae latent space using decoder
  uncertainty.
\newblock {\em Advances in Neural Information Processing Systems}, 34:802--814,
  2021.

\bibitem{tripp2020sample}
Austin Tripp, Erik Daxberger, and Jos{\'e}~Miguel Hern{\'a}ndez-Lobato.
\newblock Sample-efficient optimization in the latent space of deep generative
  models via weighted retraining.
\newblock {\em Advances in Neural Information Processing Systems},
  33:11259--11272, 2020.

\bibitem{goodfellow2014generative}
Ian Goodfellow, Jean Pouget-Abadie, Mehdi Mirza, Bing Xu, David Warde-Farley,
  Sherjil Ozair, Aaron Courville, and Yoshua Bengio.
\newblock Generative adversarial nets.
\newblock {\em Advances in neural information processing systems}, 27, 2014.

\bibitem{kingma2013auto}
Diederik~P Kingma and Max Welling.
\newblock Auto-encoding variational bayes.
\newblock {\em arXiv preprint arXiv:1312.6114}, 2013.

\bibitem{sohl2015deep}
Jascha Sohl-Dickstein, Eric Weiss, Niru Maheswaranathan, and Surya Ganguli.
\newblock Deep unsupervised learning using nonequilibrium thermodynamics.
\newblock In {\em International conference on machine learning}, pages
  2256--2265. PMLR, 2015.

\bibitem{dhariwal2021diffusion}
Prafulla Dhariwal and Alexander Nichol.
\newblock Diffusion models beat gans on image synthesis.
\newblock {\em Advances in neural information processing systems},
  34:8780--8794, 2021.

\bibitem{park2022optimization}
SM~Park, HG~Yoon, DB~Lee, JW~Choi, HY~Kwon, and C~Won.
\newblock Optimization of physical quantities in the autoencoder latent space.
\newblock {\em Scientific Reports}, 12(1):9003, 2022.

\bibitem{griffiths2020constrained}
Ryan-Rhys Griffiths and Jos{\'e}~Miguel Hern{\'a}ndez-Lobato.
\newblock Constrained bayesian optimization for automatic chemical design using
  variational autoencoders.
\newblock {\em Chemical science}, 11(2):577--586, 2020.

\bibitem{sato2023fast}
Hayaho Sato and Hajime Igarashi.
\newblock Fast topology optimization for pm motors using variational
  autoencoder and neural networks with dropout.
\newblock {\em IEEE Transactions on Magnetics}, 2023.

\bibitem{gladstone2021robust}
Rini~Jasmine Gladstone, Mohammad~Amin Nabian, Vahid Keshavarzzadeh, and Hadi
  Meidani.
\newblock Robust topology optimization using variational autoencoders.
\newblock {\em arXiv preprint arXiv:2107.10661}, 2021.

\bibitem{schumann2021machine}
Julian~F Schumann and Alejandro~M Arag{\'o}n.
\newblock A machine learning approach for fighting the curse of dimensionality
  in global optimization.
\newblock {\em arXiv preprint arXiv:2110.14985}, 2021.

\bibitem{shin2023topology}
Seungyeon Shin, Dongju Shin, and Namwoo Kang.
\newblock Topology optimization via machine learning and deep learning: A
  review.
\newblock {\em Journal of Computational Design and Engineering},
  10(4):1736--1766, 2023.

\bibitem{ramu2022survey}
Palaniappan Ramu, Pugazhenthi Thananjayan, Erdem Acar, Gamze Bayrak, Jeong~Woo
  Park, and Ikjin Lee.
\newblock A survey of machine learning techniques in structural and
  multidisciplinary optimization.
\newblock {\em Structural and Multidisciplinary Optimization}, 65(9):266, 2022.

\bibitem{nie2021topologygan}
Zhenguo Nie, Tong Lin, Haoliang Jiang, and Levent~Burak Kara.
\newblock Topologygan: Topology optimization using generative adversarial
  networks based on physical fields over the initial domain.
\newblock {\em Journal of Mechanical Design}, 143(3):031715, 2021.

\bibitem{oh2019deep}
Sangeun Oh, Yongsu Jung, Seongsin Kim, Ikjin Lee, and Namwoo Kang.
\newblock Deep generative design: Integration of topology optimization and
  generative models.
\newblock {\em Journal of Mechanical Design}, 141(11):111405, 2019.

\bibitem{maze2023diffusion}
Fran{\c{c}}ois Maz{\'e} and Faez Ahmed.
\newblock Diffusion models beat gans on topology optimization.
\newblock In {\em Proceedings of the AAAI Conference on Artificial Intelligence
  (AAAI), Washington, DC}, 2023.

\bibitem{chandrasekhar2021tounn}
Aaditya Chandrasekhar and Krishnan Suresh.
\newblock Tounn: Topology optimization using neural networks.
\newblock {\em Structural and Multidisciplinary Optimization}, 63:1135--1149,
  2021.

\bibitem{zhang2023topology}
Zeyu Zhang, Wen Yao, Yu~Li, Weien Zhou, and Xiaoqian Chen.
\newblock Topology optimization via implicit neural representations.
\newblock {\em Computer Methods in Applied Mechanics and Engineering},
  411:116052, 2023.

\bibitem{chandrasekhar2021length}
Aaditya Chandrasekhar and Krishnan Suresh.
\newblock Length scale control in topology optimization using fourier enhanced
  neural networks.
\newblock {\em arXiv preprint arXiv:2109.01861}, 2021.

\bibitem{joglekar2023dmf}
Aditya Joglekar, Hongrui Chen, and Levent~Burak Kara.
\newblock Dmf-tonn: direct mesh-free topology optimization using neural
  networks.
\newblock {\em Engineering with Computers}, pages 1--14, 2023.

\bibitem{zhong2022nsto}
Shengze Zhong, Parinya Punpongsanon, Daisuke Iwai, and Kosuke Sato.
\newblock Nsto: neural synthesizing topology optimization for modulated
  structure generation.
\newblock In {\em Computer Graphics Forum}, volume~41, pages 553--566. Wiley
  Online Library, 2022.

\bibitem{sanu2024neural}
Suryanarayanan~Manoj Sanu, Alejandro~M Aragon, and Miguel~A Bessa.
\newblock Neural topology optimization: the good, the bad, and the ugly.
\newblock {\em arXiv preprint arXiv:2407.13954}, 2024.

\bibitem{yousef2024}
Amin Yousefpour, Shirin Hosseinmardi, Carlos Mora, and Ramin Bostanabad.
\newblock Simultaneous and meshfree topology optimization with physics-informed
  gaussian processes, 2024.

\bibitem{hansen2001completely}
Nikolaus Hansen and Andreas Ostermeier.
\newblock Completely derandomized self-adaptation in evolution strategies.
\newblock {\em Evolutionary computation}, 9(2):159--195, 2001.

\bibitem{zhang2018unreasonable}
Richard Zhang, Phillip Isola, Alexei~A Efros, Eli Shechtman, and Oliver Wang.
\newblock The unreasonable effectiveness of deep features as a perceptual
  metric.
\newblock In {\em Proceedings of the IEEE conference on computer vision and
  pattern recognition}, pages 586--595, 2018.

\bibitem{pmlr-v119-fajtl20a}
Jiri Fajtl, Vasileios Argyriou, Dorothy Monekosso, and Paolo Remagnino.
\newblock Latent bernoulli autoencoder.
\newblock In Hal~Daumé III and Aarti Singh, editors, {\em Proceedings of the
  37th International Conference on Machine Learning}, volume 119 of {\em
  Proceedings of Machine Learning Research}, pages 2964--2974. PMLR, 2020.

\bibitem{larsen2016autoencoding}
Anders Boesen~Lindbo Larsen, S{\o}ren~Kaae S{\o}nderby, Hugo Larochelle, and
  Ole Winther.
\newblock Autoencoding beyond pixels using a learned similarity metric.
\newblock In {\em International conference on machine learning}, pages
  1558--1566. PMLR, 2016.

\bibitem{sosnovik2019neural}
Ivan Sosnovik and Ivan Oseledets.
\newblock Neural networks for topology optimization.
\newblock {\em Russian Journal of Numerical Analysis and Mathematical
  Modelling}, 34(4):215--223, 2019.

\bibitem{svanberg1987method}
Krister Svanberg.
\newblock The method of moving asymptotes—a new method for structural
  optimization.
\newblock {\em International journal for numerical methods in engineering},
  24(2):359--373, 1987.

\bibitem{sigmund2022benchmarking}
Ole Sigmund.
\newblock On benchmarking and good scientific practise in topology
  optimization.
\newblock {\em Structural and Multidisciplinary Optimization}, 65(11):315,
  2022.

\bibitem{evosax2022github}
Robert~Tjarko Lange.
\newblock evosax: Jax-based evolution strategies.
\newblock {\em arXiv preprint arXiv:2212.04180}, 2022.

\bibitem{hansen2009benchmarking}
Nikolaus Hansen.
\newblock Benchmarking a bi-population cma-es on the bbob-2009 function
  testbed.
\newblock In {\em Proceedings of the 11th annual conference companion on
  genetic and evolutionary computation conference: late breaking papers}, pages
  2389--2396, 2009.

\bibitem{on2021optimal}
Han-Ik On, Leekyo Jeong, Minseok Jung, Dong-Joong Kang, Jun-Hyub Park, and
  Hak-Joo Lee.
\newblock Optimal design of microwave absorber using novel variational
  autoencoder from a latent space search strategy.
\newblock {\em Materials \& Design}, 212:110266, 2021.

\bibitem{kennedy1995particle}
James Kennedy and Russell Eberhart.
\newblock Particle swarm optimization.
\newblock In {\em Proceedings of ICNN'95-international conference on neural
  networks}, volume~4, pages 1942--1948. IEEE, 1995.

\bibitem{storn1995differrential}
Rainer Storn.
\newblock Differrential evolution-a simple and efficient adaptive scheme for
  global optimization over continuous spaces.
\newblock {\em Technical report, International Computer Science Institute}, 11,
  1995.

\bibitem{lange2023discovering}
Robert Lange, Tom Schaul, Yutian Chen, Tom Zahavy, Valentin Dalibard, Chris Lu,
  Satinder Singh, and Sebastian Flennerhag.
\newblock Discovering evolution strategies via meta-black-box optimization.
\newblock In {\em Proceedings of the Companion Conference on Genetic and
  Evolutionary Computation}, pages 29--30, 2023.

\bibitem{such2017deep}
Felipe~Petroski Such, Vashisht Madhavan, Edoardo Conti, Joel Lehman, Kenneth~O
  Stanley, and Jeff Clune.
\newblock Deep neuroevolution: Genetic algorithms are a competitive alternative
  for training deep neural networks for reinforcement learning.
\newblock {\em arXiv preprint arXiv:1712.06567}, 2017.

\bibitem{liu2014efficient}
Kai Liu and Andr{\'e}s Tovar.
\newblock An efficient 3d topology optimization code written in matlab.
\newblock {\em Structural and Multidisciplinary Optimization}, 50:1175--1196,
  2014.

\bibitem{yvonnet2024topology}
Julien Yvonnet and Daicong Da.
\newblock Topology optimization to fracture resistance: a review and recent
  developments.
\newblock {\em Archives of Computational Methods in Engineering},
  31(4):2295--2315, 2024.

\bibitem{jia2023controlling}
Yingqi Jia, Oscar Lopez-Pamies, and Xiaojia~Shelly Zhang.
\newblock Controlling the fracture response of structures via topology
  optimization: From delaying fracture nucleation to maximizing toughness.
\newblock {\em Journal of the Mechanics and Physics of Solids}, 173:105227,
  2023.

\bibitem{miehe2010thermodynamically}
Christian Miehe, Fabian Welschinger, and Martina Hofacker.
\newblock Thermodynamically consistent phase-field models of fracture:
  Variational principles and multi-field fe implementations.
\newblock {\em International journal for numerical methods in engineering},
  83(10):1273--1311, 2010.

\bibitem{bourdin2008variational}
Blaise Bourdin, Gilles~A Francfort, and Jean-Jacques Marigo.
\newblock The variational approach to fracture.
\newblock {\em Journal of elasticity}, 91:5--148, 2008.

\bibitem{francfort1998revisiting}
Gilles~A Francfort and J-J Marigo.
\newblock Revisiting brittle fracture as an energy minimization problem.
\newblock {\em Journal of the Mechanics and Physics of Solids},
  46(8):1319--1342, 1998.

\bibitem{xue2023jax}
Tianju Xue, Shuheng Liao, Zhengtao Gan, Chanwook Park, Xiaoyu Xie, Wing~Kam
  Liu, and Jian Cao.
\newblock Jax-fem: A differentiable gpu-accelerated 3d finite element solver
  for automatic inverse design and mechanistic data science.
\newblock {\em Computer Physics Communications}, page 108802, 2023.

\bibitem{jax2018github}
James Bradbury, Roy Frostig, Peter Hawkins, Matthew~James Johnson, Chris Leary,
  Dougal Maclaurin, George Necula, Adam Paszke, Jake Vander{P}las, Skye
  Wanderman-{M}ilne, and Qiao Zhang.
\newblock {JAX}: composable transformations of {P}ython+{N}um{P}y programs,
  2018.

\bibitem{ulyanov2018deep}
Dmitry Ulyanov, Andrea Vedaldi, and Victor Lempitsky.
\newblock Deep image prior.
\newblock In {\em Proceedings of the IEEE conference on computer vision and
  pattern recognition}, pages 9446--9454, 2018.

\bibitem{rahaman2019spectral}
Nasim Rahaman, Aristide Baratin, Devansh Arpit, Felix Draxler, Min Lin, Fred
  Hamprecht, Yoshua Bengio, and Aaron Courville.
\newblock On the spectral bias of neural networks.
\newblock In {\em International Conference on Machine Learning}, pages
  5301--5310. PMLR, 2019.

\end{thebibliography}

\newpage
\appendix

\begin{appendix}

\section{Complementary results}

\begin{figure*}[h]
    \centering
    \includegraphics[width =0.9\textwidth]{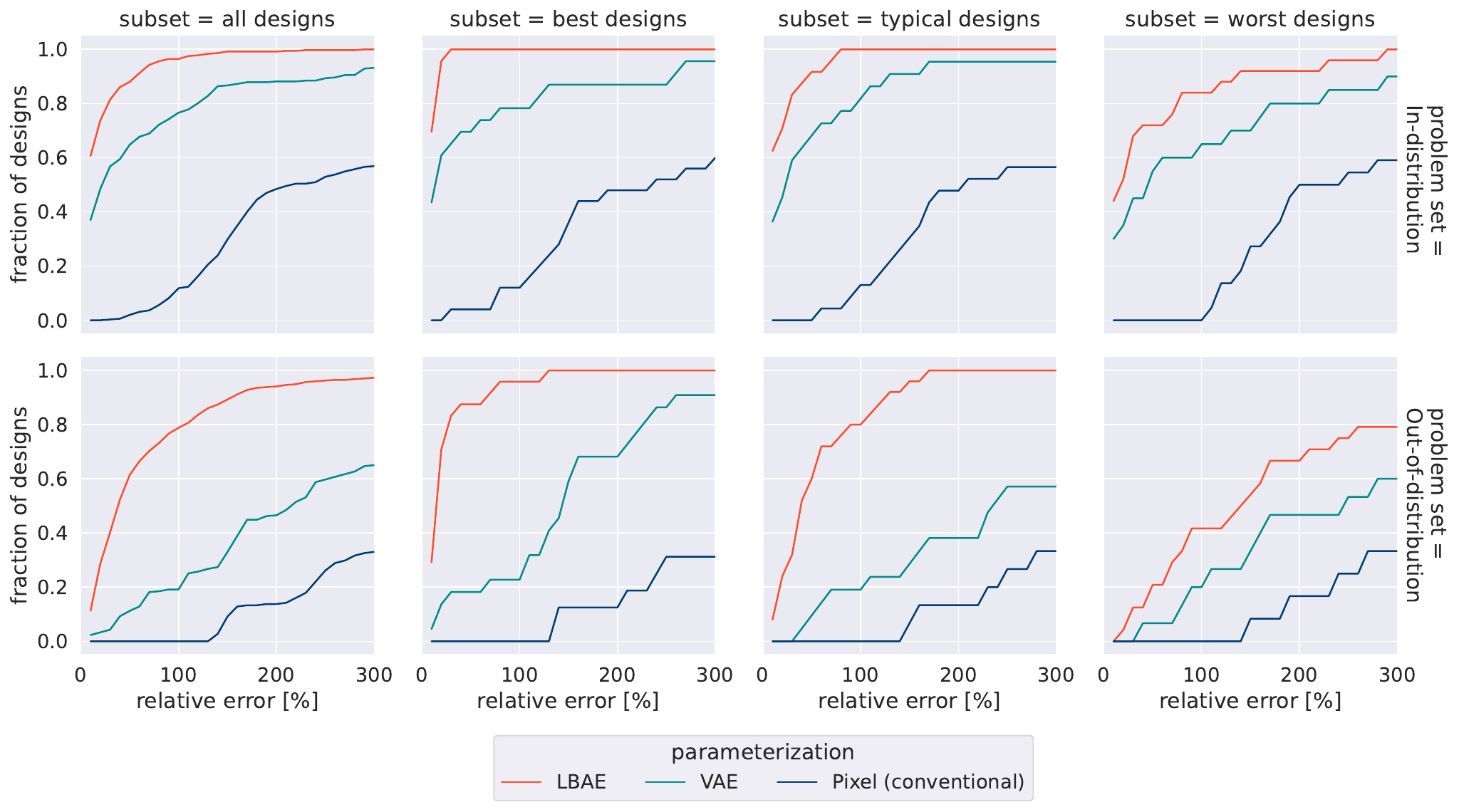}
    \caption{Results for different reparameterization models, optimized with CMA-ES optimizer with default settings.}
    \label{fig:a1}
\end{figure*}

\begin{figure*}[h]
    \centering
    \includegraphics[width = 0.9\textwidth]{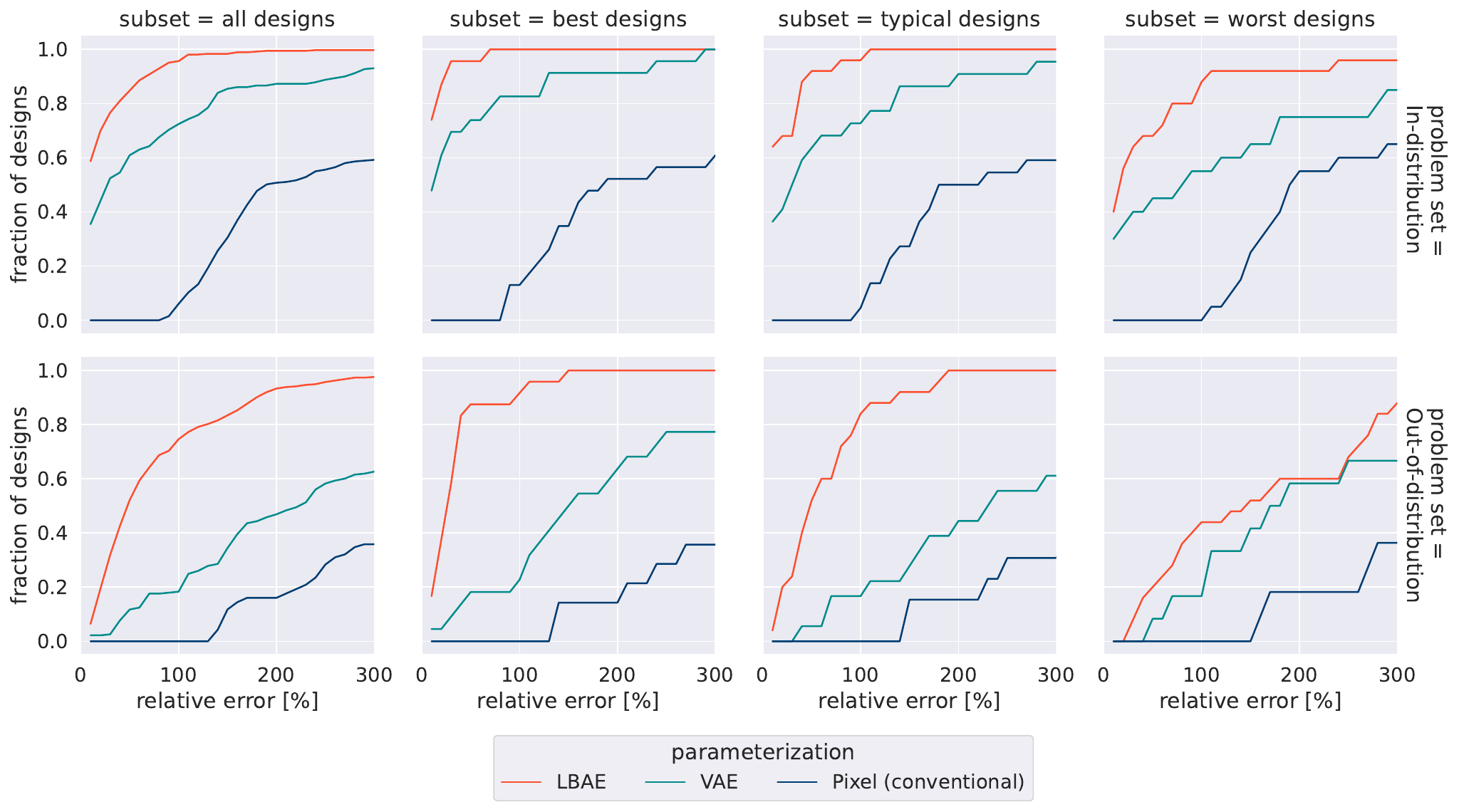}
    \caption{Results for different reparameterization models, optimized with CMA-ES optimizer with customized settings.}
    \label{fig:a2}
\end{figure*}

\begin{figure*}[h]
    \centering
    \includegraphics[width =0.9\textwidth]{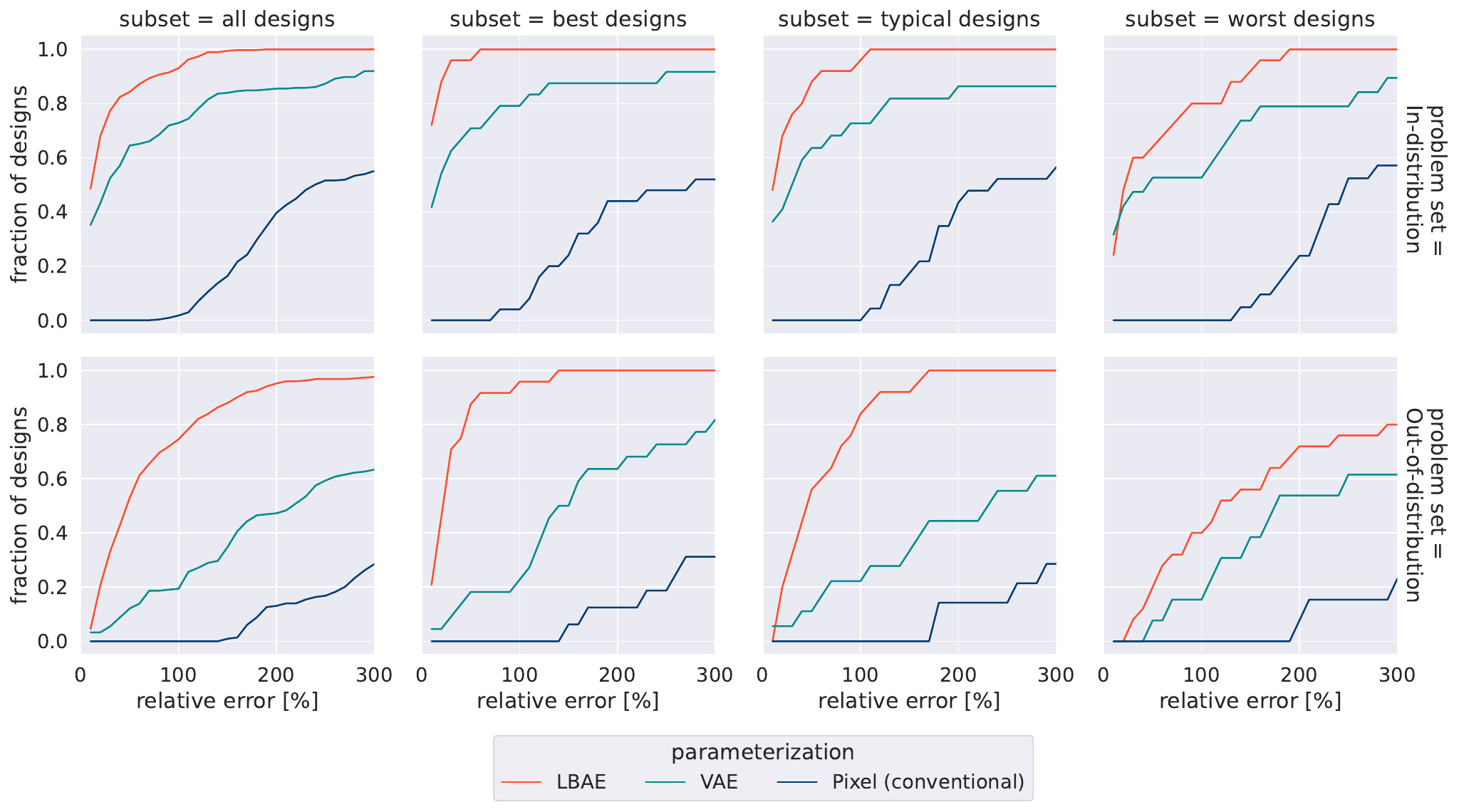}
    \caption{Results for different reparameterization models, optimized with BIPOP-CMA-ES optimizer with customized settings.}
    \label{fig:a3}
\end{figure*}

Figures \ref{fig:a1}, \ref{fig:a2}, and \ref{fig:a3} present results for the remaining optimizer configurations: CMA-ES with default settings, as well as CMA-ES and BIPOP-CMA-ES with customized settings (added variable clipping and initialization, both within the range of [-5, 5]).

\section{Thresholded designs}

\begin{figure*}[h]
    \centering
    \includegraphics[width =0.9 \textwidth]{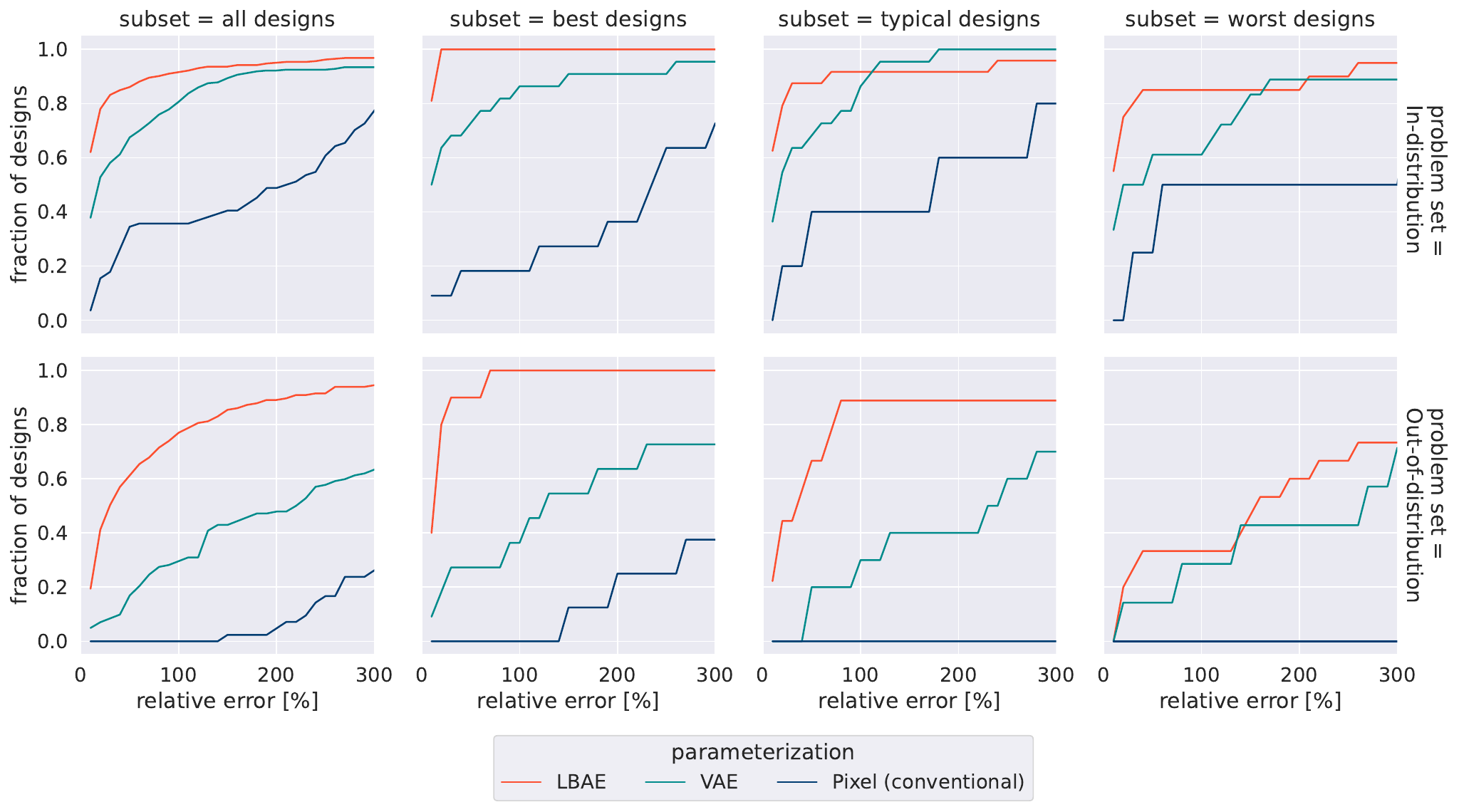}
    \caption{Performance of the thresholded designs, optimized with CMA-ES with default settings.}
    \label{fig:b1}
\end{figure*}

\begin{figure*}[h]
    \centering
    \includegraphics[width =0.9 \textwidth]{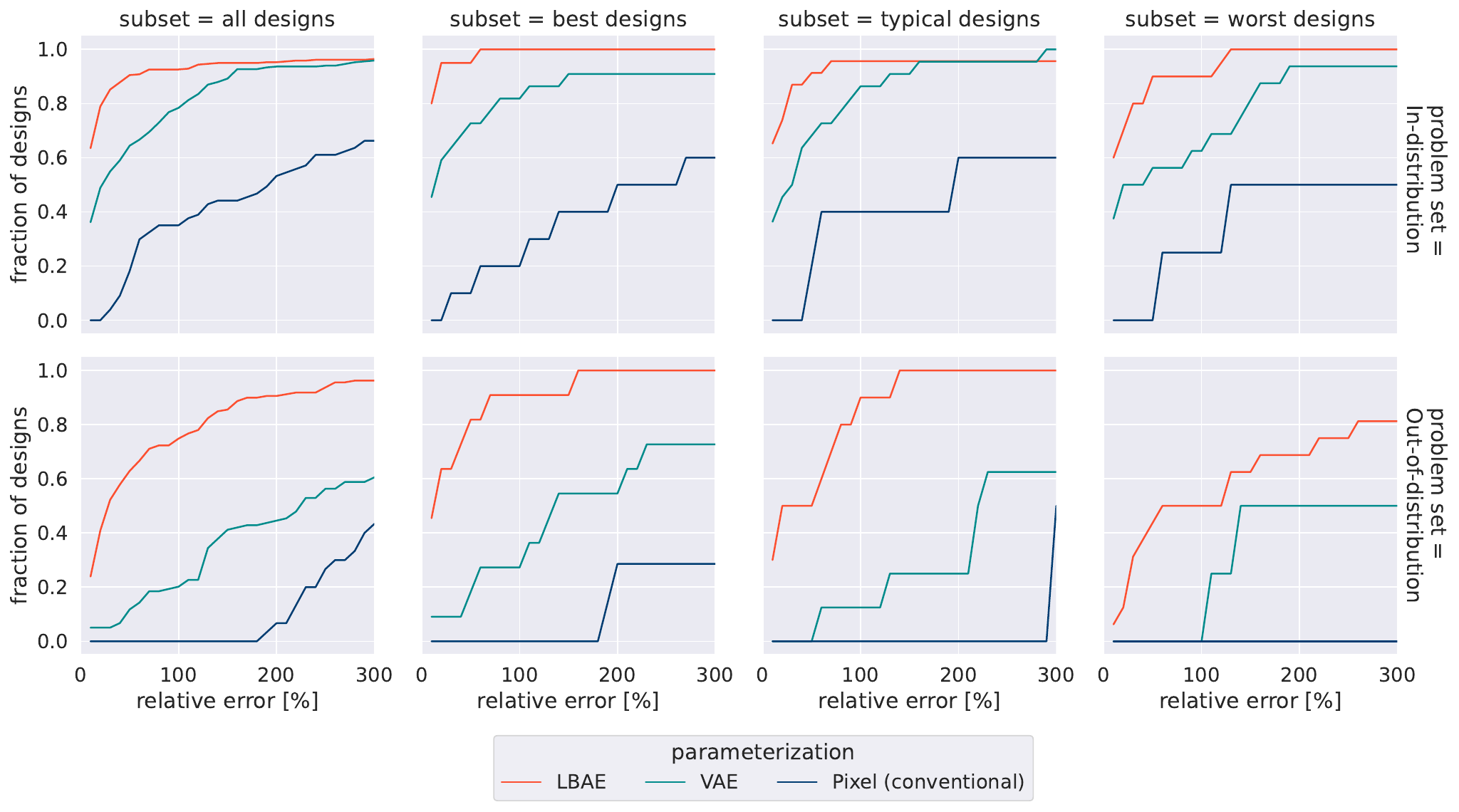}
    \caption{Performance of the thresholded designs, optimized with CMA-ES with custom settings (variable clipping and initialization within the range [-5, 5]).}
    \label{fig:b2}
\end{figure*}

\begin{figure*}[h]
    \centering
    \includegraphics[width = 0.9\textwidth]{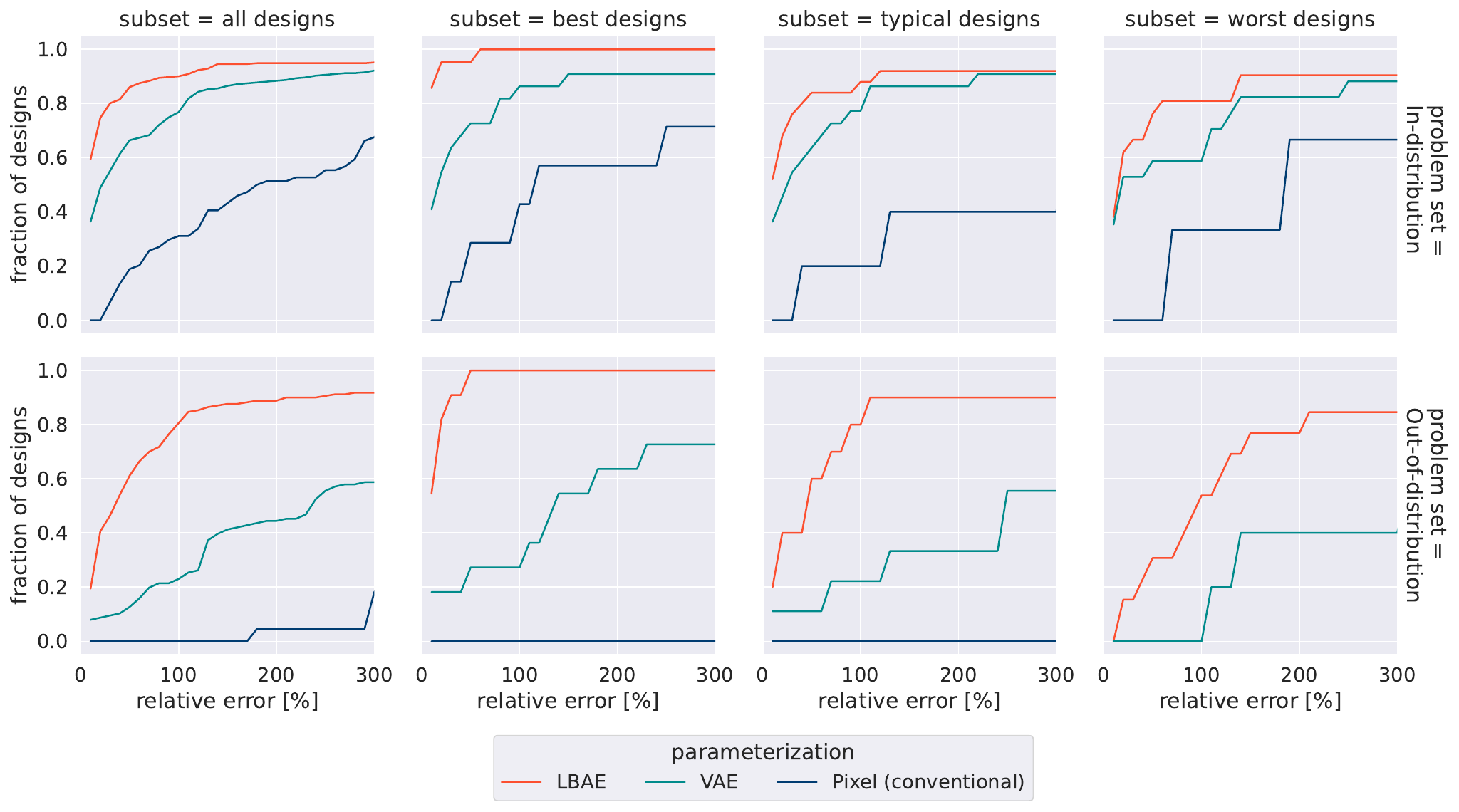}
     \caption{Performance of the thresholded designs, optimized with BIPOP-CMA-ES with custom settings (variable clipping and initialization within the range [-5, 5]).}
    \label{fig:b3}
\end{figure*}

Figures \ref{fig:b1}, \ref{fig:b2}, and \ref{fig:b3} present the results for thresholded designs. The relative error is calculated with respect to the compliance obtained for a thresholded MMA design.

\subsection{Comparison of optimizers}

\begin{figure*}[h]
    \centering
    \includegraphics[width =\textwidth]{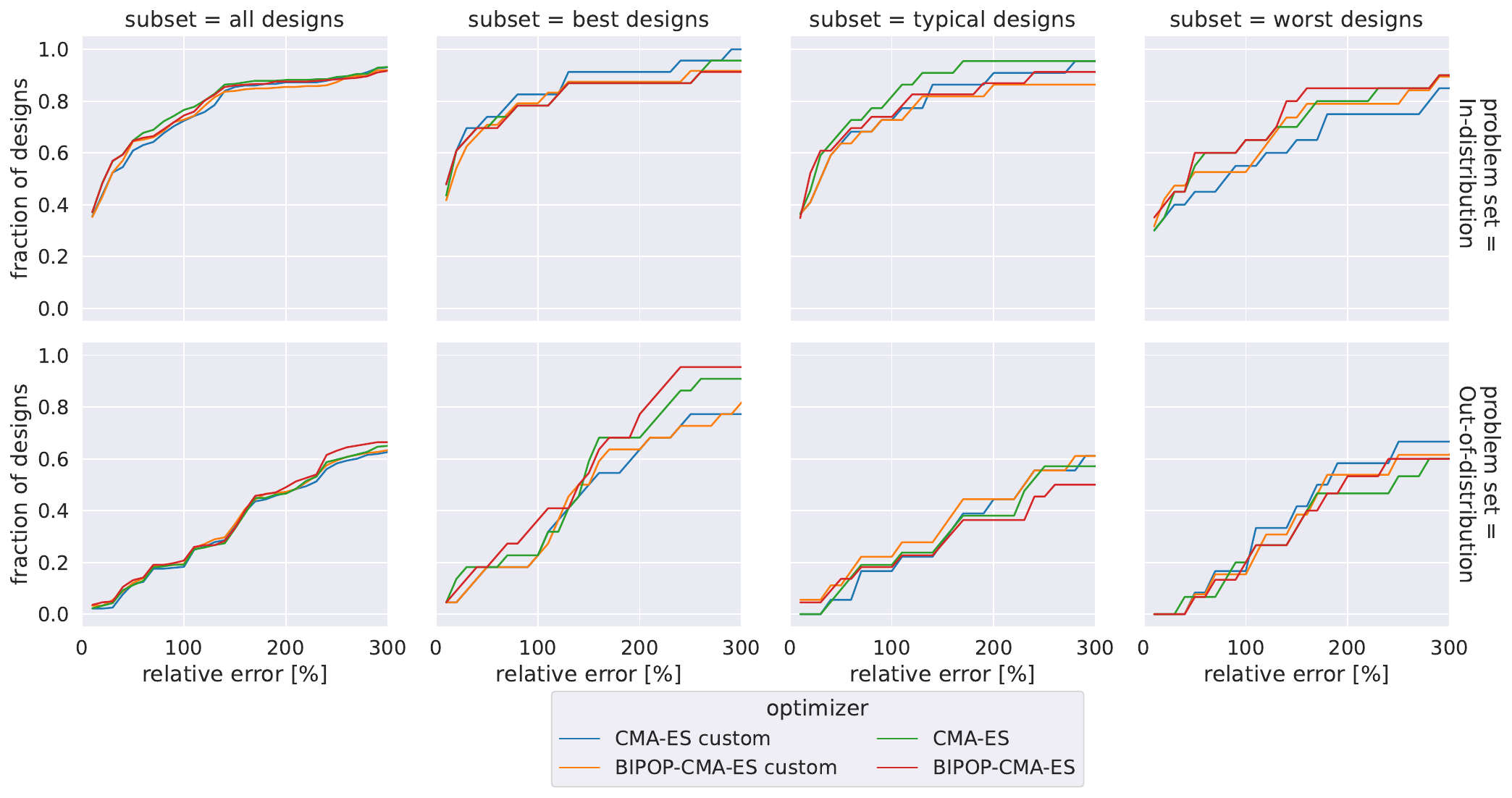}
    \caption{Comparison of optimizers using VAE parameterization.}
    \label{fig:optim1}
\end{figure*}

\begin{figure*}[h]
    \centering
    \includegraphics[width = \textwidth]{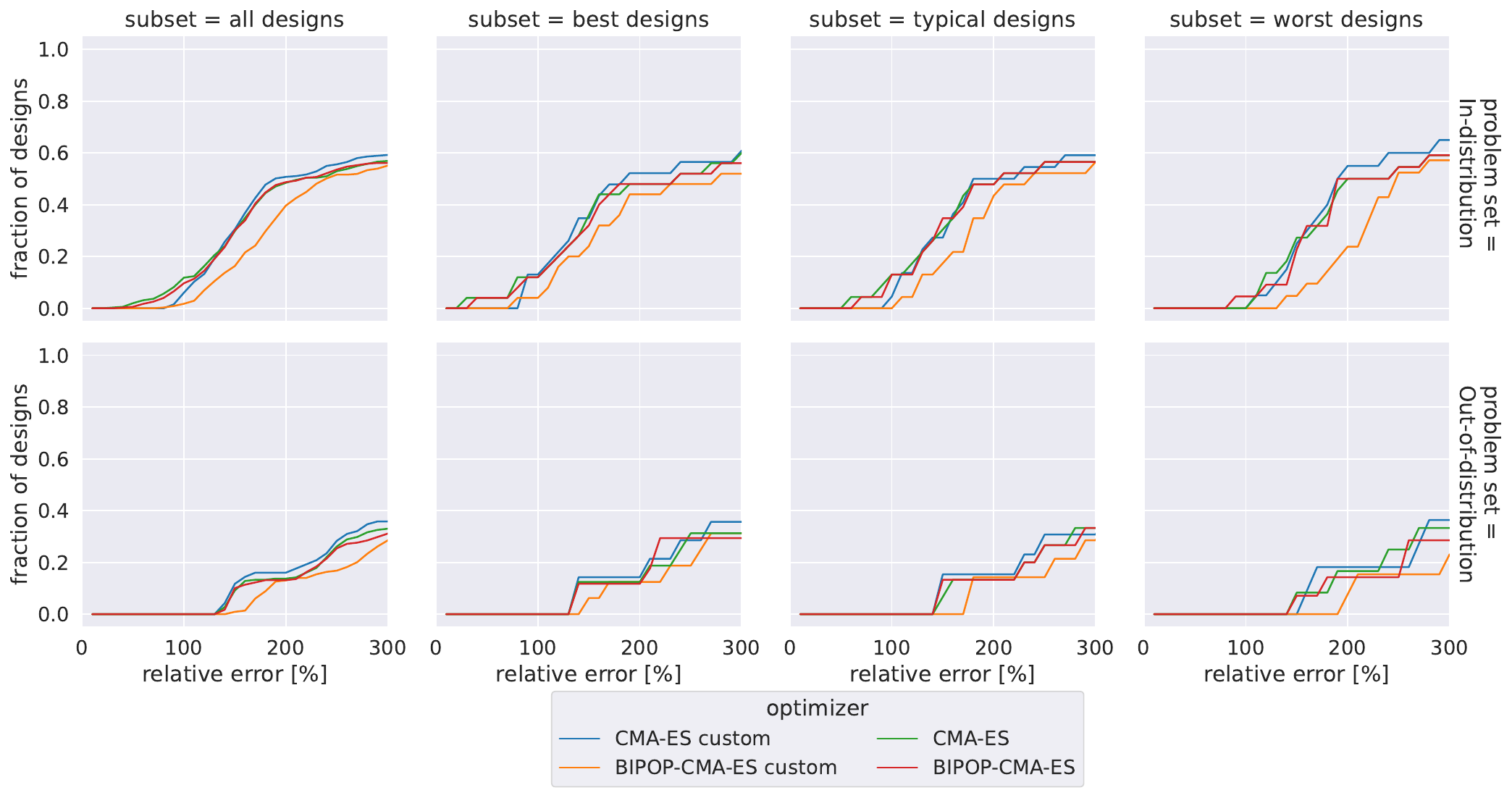}
    \caption{Comparison of optimizers using conventional pixel parameterization.}
    \label{fig:optim2}
\end{figure*}

\begin{figure*}[h]
    \centering
    \includegraphics[width = 0\textwidth]{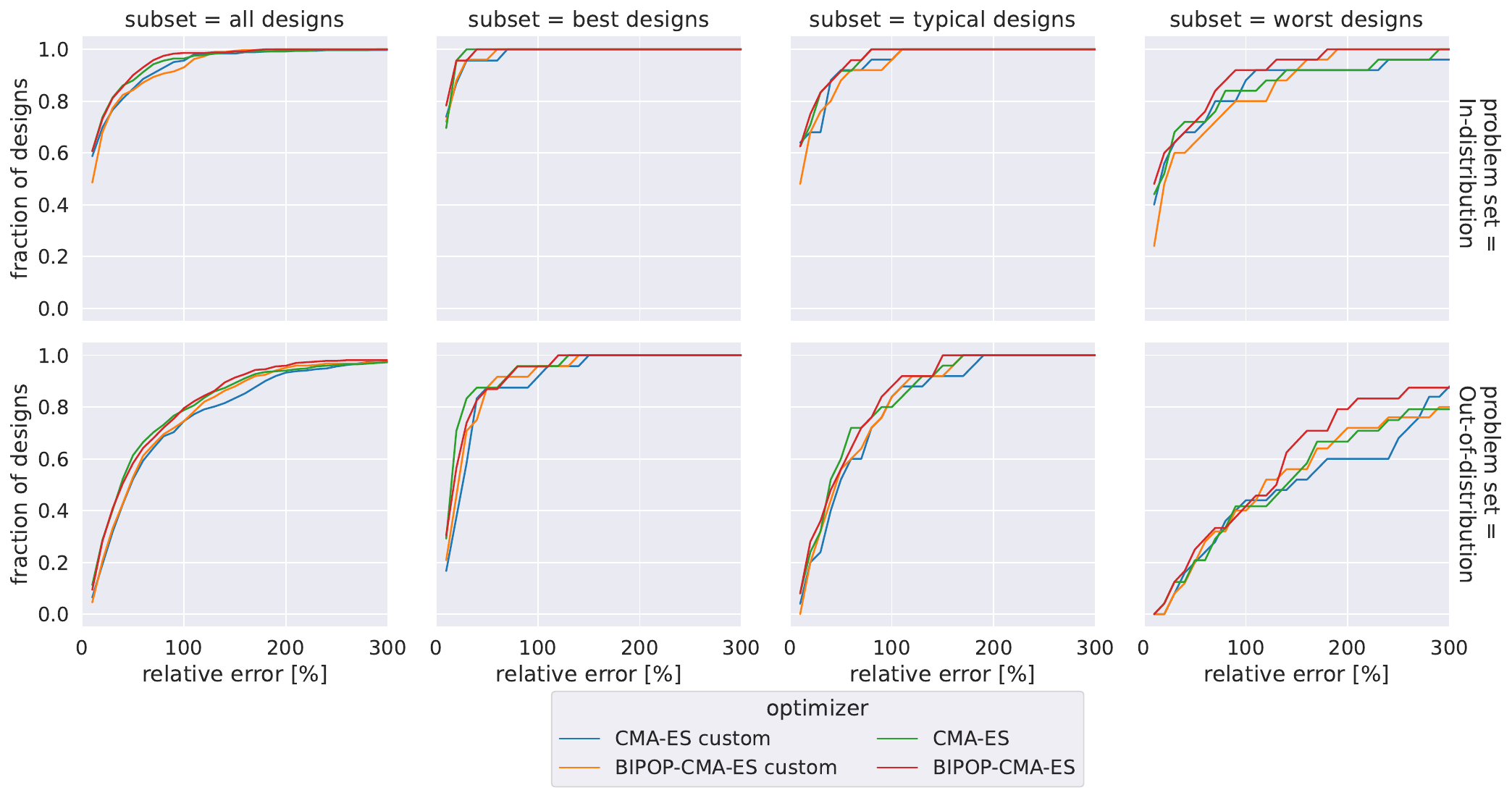}
    \caption{Comparison of optimizers using LBAE parameterization.}
    \label{fig:optim3}
\end{figure*}

Figures \ref{fig:optim1}, \ref{fig:optim2}, and \ref{fig:optim3} compare the performance of two different optimizers, with different settings (default and custom). In the custom configuration, we apply clipping of the latent variables in the range of [-5, 5], and initialize the latent variable within that range. 

\subsection{Optimization with thresholding}

\begin{figure*}[h]
    \centering
    \includegraphics[width = \textwidth]{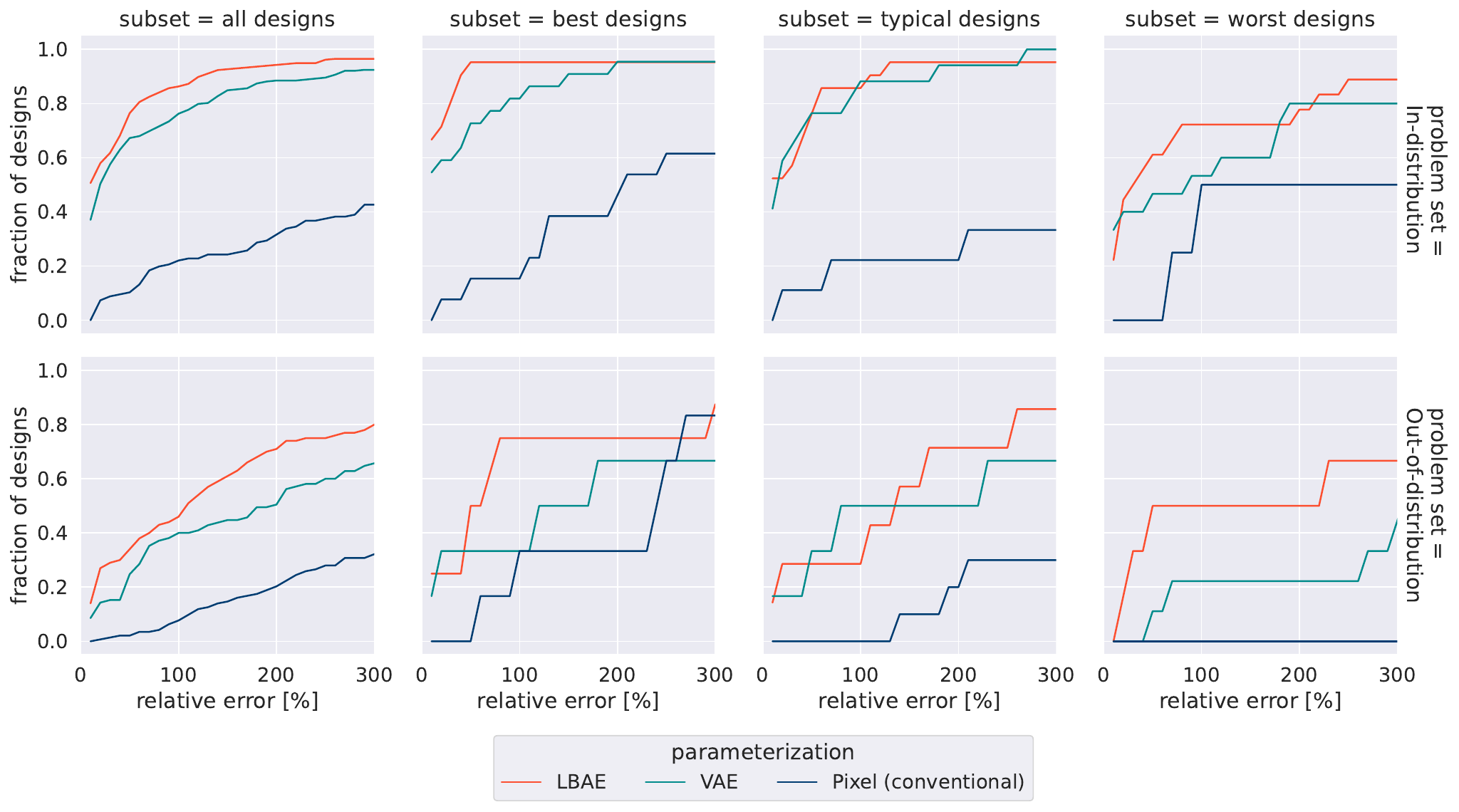}
    \caption{Performance of designs optimized with CMA-ES, using thresholding during optimization. }
    \label{fig:th2}
\end{figure*}

The results shown in Figure \ref{fig:th2} were obtained using thresholding during optimization, i.e. the objective value was evaluated and optimized on an already thresholded sample design. 

\subsection{Correction of the latent vector}

\begin{figure*}[h]
    \centering
    \includegraphics[width = \textwidth]{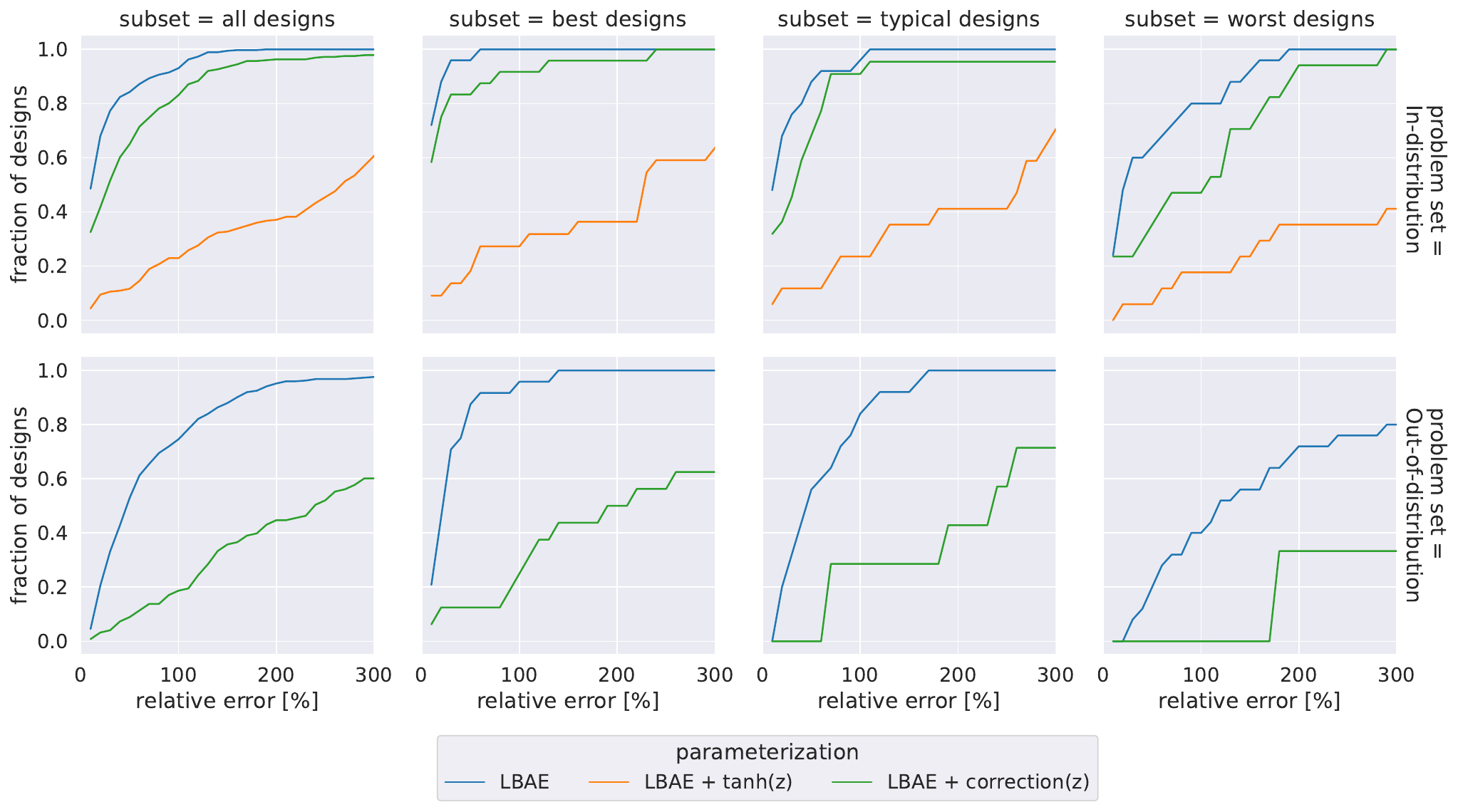}
    \caption{Results for different correction schemes, compared to the default LBAE configuration, optimized using BIPOP-CMA-ES optimizer.}
    \label{fig:cors1}
\end{figure*}

To account for the limitations arising from imposing the volume constraint in a hard way, we experiment with a strategy to account for that effect by correcting the latent representation. Given a latent representation, we generate a target design with an imposed volume fraction constraint (sigmoid with a constrained mean). Given the same latent representation, we generate a design without the constraint (applying only standard sigmoid activation in the output layer). The latent vector is then optimized by minimizing the MSE loss between the target design (with the volume constraint) and the design without the constraint. Gradients are backpropagated through the decoder network, and updates to the latent vector are applied using the ADAM optimizer, with a learning rate of 0.1, using 10 iterations. The results are compared with the standard approach, without correction in Figure \ref{fig:cors1}.

Additionally, we experiment with applying tanh activation on the latent variable, before passing it through the decoder network, as a way to emulate the Bernoulli distribution, as experienced during the training. As can be seen in Figure \ref{fig:cors1}, both strategies turn out to deteriorate the performance.

\subsection{Convergence rate estimations}

\begin{figure*}[h]
    \centering
    \includegraphics[width = 0.9\textwidth]{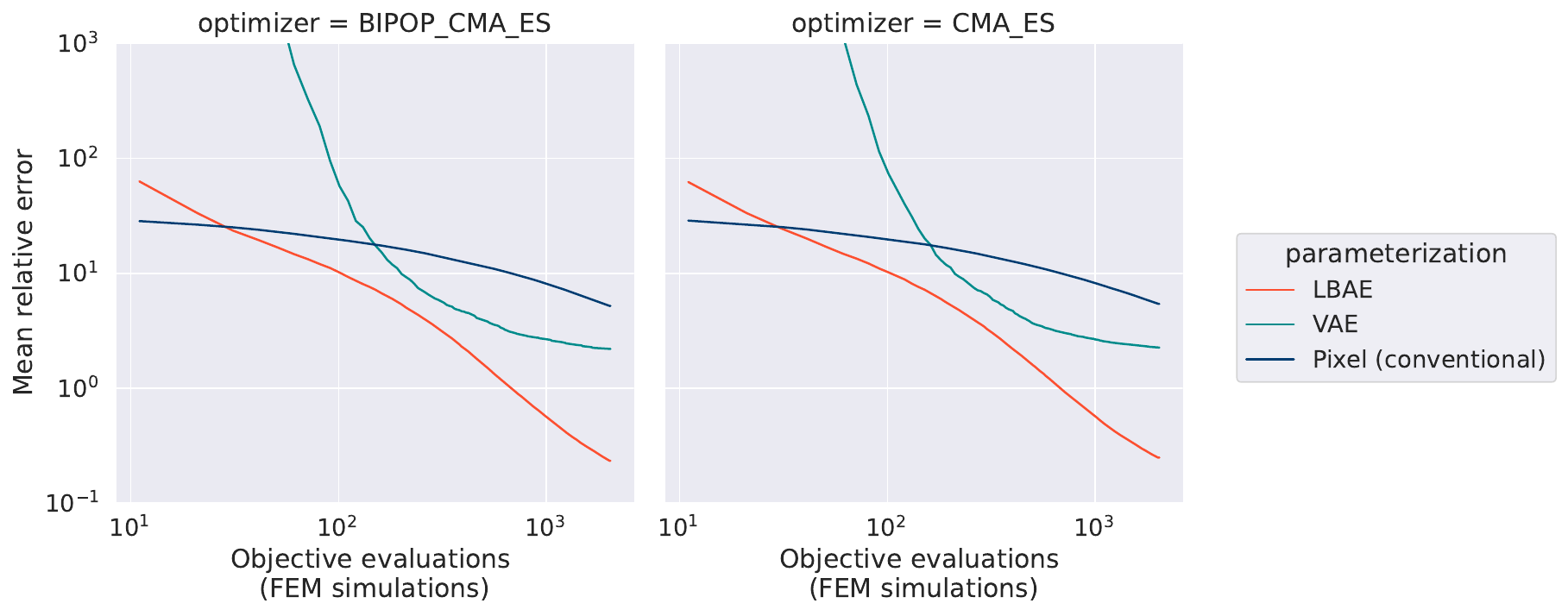}
    \caption{Results for different correction schemes, compared to the default LBAE configuration, optimized using BIPOP-CMA-ES optimizer.}
    \label{fig:cors2}
\end{figure*}

Figure \ref{fig:cors2} presents the mean error (averaged over all designs from both datasets, disregarding 10\% of the worst runs) vs the number of objective evaluations. Although each problem is characterized by a different optimization history, affected e.g. by initialization, the plot gives a qualitative indication of the convergence rates between the 3 different parameterization schemes. 

\section{Shape similarity argument}

\begin{figure*}
    \centering
    \includegraphics[width = \textwidth]{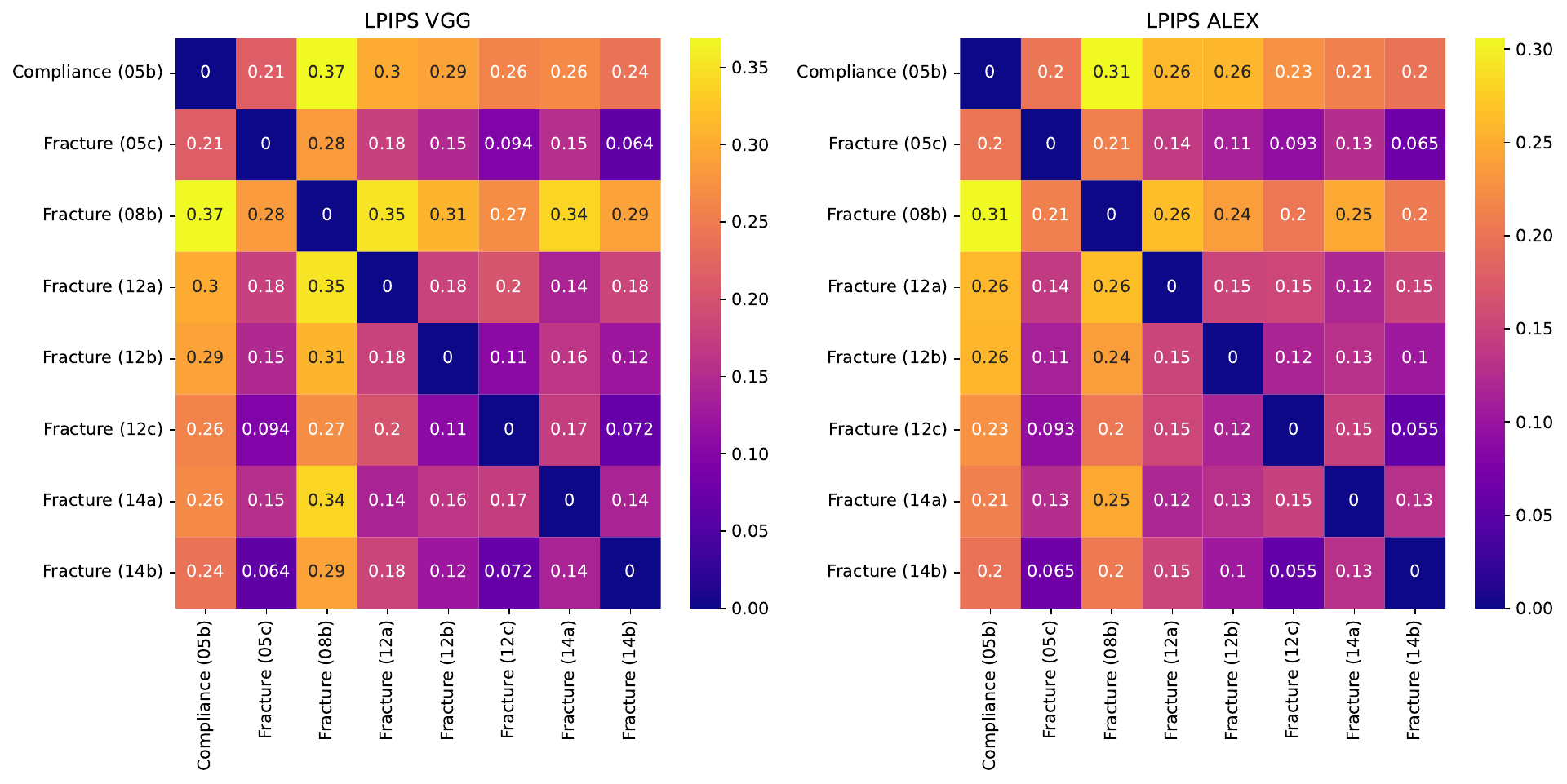}
    \caption{Learned Perceptual Image Patch Similarity (LPIPS) index, quantifying similarity between the designs reported by Desai et al \cite{desai2022topology} optimized for compliance and for fracture using different hyperparameters.}
    \label{fig:shape-sim}
\end{figure*}

Here, we present a quantitative argument supporting our assumption that design similarity could be a reasonable proxy for good performance. In other words, we show that designs optimized for different objectives oftentimes are similar (share the same features). We consider the designs of the cantilever reported in the work by Desai et al. \cite{desai2022topology}, and compare the design optimized for compliance with the designs optimized for fracture with different hyperparameter settings. We quantify similarity using the Learned Perceptual Image Patch Similarity (LPIPS) metric \cite{zhang2018unreasonable}, which provides superior performance to quantify perceptual similarity using the feature maps of pre-trained neural networks. The similarity indices for different designs are shown in Figure \ref{fig:shape-sim}, calculated using two different feature map extractors (VGG and Alex-Net). 

The compliance design is slightly more dissimilar to the fracture designs - as indicated by a higher LPIPS index. However, we also find a counterexample, the fracture design (extracted from Figure 8b in \cite{desai2022topology}), with a comparably high similarity index. In other words, the fracture design (8b) is as similar to the remaining fracture designs as the compliance design. The findings are consistent regardless of which feature extractor was used for the LPIPS calculation.

\section{Latent dimensionality and population size trade-off}
\label{app:d}
Table \ref{fig:tradeoff} illustrates the trade-off present while choosing the dimensionality of the latent space. Models with larger latent dimensionality are easier to train (requiring less compression). On the other hand, larger dimensionality hinders the optimization performance of gradient-free optimizers. In our study, the latent dimensionality of 256 was found to balance these two factors best.

\begin{table}
\centering
\caption{Average relative error (wrt. MMA solution) obtained with LBAE models with varying latent space dimensionality, and with different population sizes (CMA-ES).}
\label{fig:tradeoff}
\begin{tabular}{@{}lllll@{}}
\toprule
 & \multicolumn{4}{c}{Population size} \\ \midrule
\begin{tabular}[c]{@{}l@{}}Latent dimension\end{tabular} & 4 & 8 & 16 & 32 \\\midrule
512 & 1.156 & 0.575 & 0.600 & 0.712 \\
256 & 1.289 & \textbf{0.507} & 0.512 & 0.659 \\
128 & 3.637 & 0.817 & 0.623 & 0.761 \\
64 & 15.677 & 1.151 & 0.835 & 1.090 \\
32 & 2.777 & 1.709 & 1.280 & 1.642 \\ \bottomrule
\end{tabular}
\end{table}

Similarly, the optimal population size (number of samples for an update of the gradient-free optimizer), is also governed by a trade-off. In this case, it is the frequency of the updates (lower population -- more frequent updates) compared to the quality of the updates (higher population -- better estimation of the optimal update). In that regard, the population size on the order of 10 samples provides the best performance. 

\section{Model Architectures}
\label{app:E}
The details of the LBAE model architectures are listed in Table \ref{tab:lbae-encoder} and \ref{tab:lbae-decoder}.

\begin{table}[h!]
\centering
\caption{LBAE Encoder Architecture. In each residual block, the residual connection is defined between the output of the first and last batch norm.}\label{tab:lbae-encoder}
\begin{tabular}{|l|}
\hline
\textbf{Input features block} \\ \hline
Conv2d(1, 64, kernel\_size=(3, 3), stride=(1, 1), padding=(1, 1), bias=False)    \\ 
BatchNorm2d(64, eps=1e-05, momentum=0.1, affine=True, track\_running\_stats=True) \\ 
LeakyReLU(negative\_slope=0.2)                                                   \\ \hline
\multicolumn{1}{|c|}{\textbf{Residual Blocks (3)}}                              \\ \hline
\multicolumn{1}{|l|}{\textbf{Block 1}}                                           \\ \hline

Conv2d(64, 64, kernel\_size=(4, 4), stride=(2, 2), padding=(1, 1), bias=False)   \\ 
BatchNorm2d(64, eps=1e-05, momentum=0.1, affine=True, track\_running\_stats=True) \\ 
Conv2d(64, 64, kernel\_size=(3, 3), stride=(1, 1), padding=(1, 1), bias=False)   \\ 
BatchNorm2d(64, eps=1e-05, momentum=0.1, affine=True, track\_running\_stats=True) \\ 
Conv2d(64, 64, kernel\_size=(3, 3), stride=(1, 1), padding=(1, 1), bias=False)   \\ 
BatchNorm2d(64, eps=1e-05, momentum=0.1, affine=True, track\_running\_stats=True) \\ 
LeakyReLU(negative\_slope=0.2)                                                   \\ \hline
\multicolumn{1}{|l|}{\textbf{Block 2}}                                           \\ \hline
Conv2d(64, 128, kernel\_size=(4, 4), stride=(2, 2), padding=(1, 1), bias=False)  \\ 
BatchNorm2d(128, eps=1e-05, momentum=0.1, affine=True, track\_running\_stats=True)\\ 
Conv2d(128, 128, kernel\_size=(3, 3), stride=(1, 1), padding=(1, 1), bias=False) \\ 
BatchNorm2d(128, eps=1e-05, momentum=0.1, affine=True, track\_running\_stats=True)\\ 
Conv2d(128, 128, kernel\_size=(3, 3), stride=(1, 1), padding=(1, 1), bias=False) \\ 
BatchNorm2d(128, eps=1e-05, momentum=0.1, affine=True, track\_running\_stats=True)\\ 
LeakyReLU(negative\_slope=0.2)                                                   \\ \hline
\multicolumn{1}{|l|}{\textbf{Block 3}}                                           \\ \hline
Conv2d(128, 256, kernel\_size=(4, 4), stride=(2, 2), padding=(1, 1), bias=False) \\ 
BatchNorm2d(256, eps=1e-05, momentum=0.1, affine=True, track\_running\_stats=True)\\ 
Conv2d(256, 256, kernel\_size=(3, 3), stride=(1, 1), padding=(1, 1), bias=False) \\ 
BatchNorm2d(256, eps=1e-05, momentum=0.1, affine=True, track\_running\_stats=True)\\ 
Conv2d(256, 256, kernel\_size=(3, 3), stride=(1, 1), padding=(1, 1), bias=False) \\ 
BatchNorm2d(256, eps=1e-05, momentum=0.1, affine=True, track\_running\_stats=True)\\ \hline
\multicolumn{1}{|l|}{\textbf{Latent features block}}   \\ \hline
Conv2d(256, 256, kernel\_size=(4, 4), stride=(2, 2), padding=(1, 1), bias=False) \\ 
LeakyReLU(negative\_slope=0.2)                                                   \\ 
Linear(in\_features=4096, out\_features=256, bias=True)                          \\
Tanh() \\ \hline
\end{tabular}
\end{table}

\begin{table}[h!]
\centering
\caption{LBAE Decoder Architecture. In each residual block, the residual connection is defined between the output of the first and last batch norm.}\label{tab:lbae-decoder}
\begin{tabular}{|l|}
\hline
\textbf{Latent features block} \\ \hline
Linear(in\_features=256, out\_features=4096, bias=True)                          \\ 
LeakyReLU(negative\_slope=0.2)                                                   \\ \hline
\multicolumn{1}{|c|}{\textbf{Residual Blocks (3)}}                              \\ \hline
\multicolumn{1}{|l|}{\textbf{Block 1}}                                           \\ \hline
ConvTranspose2d(256, 256, kernel\_size=(4, 4), stride=(2, 2), padding=(1, 1), bias=False) \\ 
BatchNorm2d(256, eps=1e-05, momentum=0.1, affine=True, track\_running\_stats=True) \\ 
ConvTranspose2d(256, 256, kernel\_size=(3, 3), stride=(1, 1), padding=(1, 1), bias=False) \\ 
BatchNorm2d(256, eps=1e-05, momentum=0.1, affine=True, track\_running\_stats=True) \\ 
ConvTranspose2d(256, 256, kernel\_size=(3, 3), stride=(1, 1), padding=(1, 1), bias=False) \\ 
BatchNorm2d(256, eps=1e-05, momentum=0.1, affine=True, track\_running\_stats=True) \\ 
LeakyReLU(negative\_slope=0.2)                                                   \\ \hline
\multicolumn{1}{|l|}{\textbf{Block 2}}                                           \\ \hline
ConvTranspose2d(256, 128, kernel\_size=(4, 4), stride=(2, 2), padding=(1, 1), bias=False) \\ 
BatchNorm2d(128, eps=1e-05, momentum=0.1, affine=True, track\_running\_stats=True) \\ 
ConvTranspose2d(128, 128, kernel\_size=(3, 3), stride=(1, 1), padding=(1, 1), bias=False) \\ 
BatchNorm2d(128, eps=1e-05, momentum=0.1, affine=True, track\_running\_stats=True) \\ 
ConvTranspose2d(128, 128, kernel\_size=(3, 3), stride=(1, 1), padding=(1, 1), bias=False) \\ 
BatchNorm2d(128, eps=1e-05, momentum=0.1, affine=True, track\_running\_stats=True) \\ 
LeakyReLU(negative\_slope=0.2)                                                   \\ \hline
\multicolumn{1}{|l|}{\textbf{Block 3}}                                           \\ \hline
ConvTranspose2d(128, 64, kernel\_size=(4, 4), stride=(2, 2), padding=(1, 1), bias=False) \\ 
BatchNorm2d(64, eps=1e-05, momentum=0.1, affine=True, track\_running\_stats=True) \\ 
ConvTranspose2d(64, 64, kernel\_size=(3, 3), stride=(1, 1), padding=(1, 1), bias=False) \\ 
BatchNorm2d(64, eps=1e-05, momentum=0.1, affine=True, track\_running\_stats=True) \\ 
ConvTranspose2d(64, 64, kernel\_size=(3, 3), stride=(1, 1), padding=(1, 1), bias=False) \\ 
BatchNorm2d(64, eps=1e-05, momentum=0.1, affine=True, track\_running\_stats=True) \\ 
LeakyReLU(negative\_slope=0.2)                                                   \\ \hline
\multicolumn{1}{|l|}{\textbf{Output block}}   \\ \hline
ConvTranspose2d(64, 64, kernel\_size=(4, 4), stride=(2, 2), padding=(1, 1), bias=False) \\ 
BatchNorm2d(64, eps=1e-05, momentum=0.1, affine=True, track\_running\_stats=True) \\ 
ConvTranspose2d(64, 1, kernel\_size=(3, 3), stride=(1, 1), padding=(1, 1), bias=False) \\ \hline
\end{tabular}
\end{table}

\subsection{Alternative configurations of the baseline VAE model}
For the baseline VAE model, we tested 4 different configurations: 2 different architecture configurations (see Table \ref{tab1}), and 2 different parameterizations of the latent space. The results are summarized in Figure \ref{fig:stats2}. 

\begin{figure*}
  \centering
  \includegraphics[width= 1.0\textwidth]{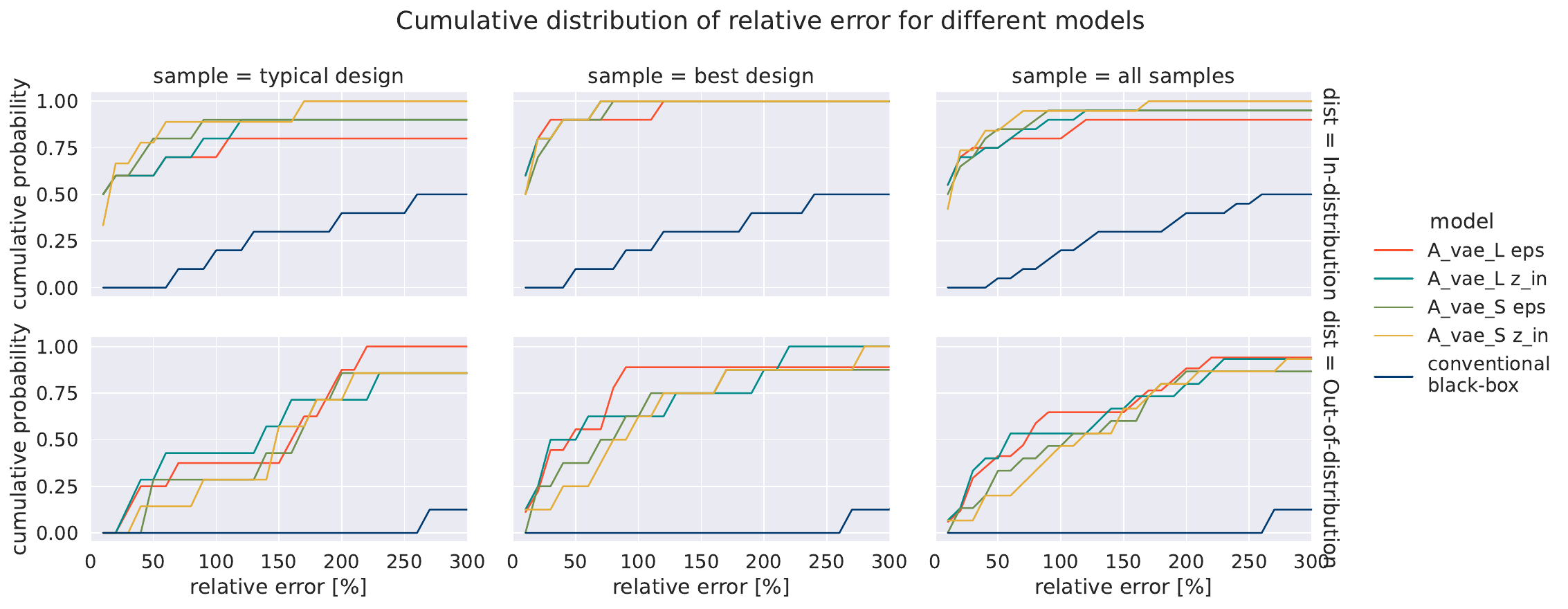}
  \caption{Cumulative probability vs relative error (w.r.t. the MMA solution) for different optimization methods, evaluated on smaller testing sets (10 problems each, unlike on the full results), repeated with 10 different random seeds. The distribution does not add up to 1 due to cut-off at 300\%.}
  \label{fig:stats26}
\end{figure*}

\begin{table}
  \centering
  \caption{Two VAE architecture configurations were used in this study. The decoder uses the same layers but in reversed order, with bilinear upsampling in between the convolutional layers to increase the resolution.}
  \label{tab1}
  \begin{tabular}{@{}ll@{}}
  \toprule

  \multicolumn{1}{l}{Encoder configuration L} & {Encoder configuration S} \\ \midrule
  \multicolumn{2}{l}{convolutional layers (4)} \\ \midrule
  \multicolumn{1}{l}{\begin{tabular}[c]{@{}l@{}}filters: 64, 64, 128, 128\\ stride 2\\ kernel size: 4\\ padding: 1\\ activation: Leaky ReLU\end{tabular}} & \begin{tabular}[c]{@{}l@{}}filters: 32, 32, 64, 64\\ stride 2\\ kernel size: 4\\ padding 1\\ activation: Leaky ReLU\end{tabular} \\ \midrule
  \multicolumn{2}{l}{Dense layer:} \\ \midrule
  \multicolumn{1}{l}{\begin{tabular}[c]{@{}l@{}}input dim: 2048,\\ output dim: 512,\\ activation: Leaky ReLU\end{tabular}} & \begin{tabular}[c]{@{}l@{}}input dim: 1024,\\ output dim: 256,\\ activation: Leaky ReLU\end{tabular} \\ \midrule
  \multicolumn{2}{l}{Dense layer:} \\ \midrule
  \multicolumn{1}{l}{\begin{tabular}[c]{@{}l@{}}input dim: 512\\ output dim: 2x64 ($\mu$ and $\sigma$)\\ activation: linear\end{tabular}} & \begin{tabular}[c]{@{}l@{}}input dim: 256\\ output dim: 2x32 ($\mu$ and $\sigma$)\\ activation: linear\end{tabular} \\ \bottomrule
  \end{tabular}
  \end{table}

\paragraph{Parametrization}

For optimization with VAE, we explore two different parameterizations of the latent space: 1) where the latent optimizer controls the entire input to the decoder ($\vec{z}$), and 2) where we utilize the information from the encoder, and the optimizer controls the additive term $\epsilon$, such that the input to the decoder $z = \mu + \sigma \epsilon$, where $\mu$ and $\sigma$ are outputs of the encoder. In this second approach, we feed the best design obtained so far throughout the optimization roll-out, initially starting with a uniformly distributed material. In such a way, the optimizer controls the perturbation of the best design so far, rather than the complete design. In the results presented in the article, we use the larger architecture (VAE-L) with standard parameterization of the latent space (i.e., controlling the full latent vector).

\section{Phase-field fracture simulation}

\begin{table}[h]
\caption{Detailed parameters of the phase-fields fracture simulation}
\centering
\begin{tabular}{|l|l|}
\hline
Parameter & value \\ \hline
Young's modulus: E & 210 GPa \\ \hline
Critical energy release rate: $G_c$ & 2.7 kJ/$m^2$ \\ \hline
Poisson ratio: $\nu$ & 0.3 \\ \hline
Phase field characteristic length scale: l & 2 mm \\ \hline
Staggered scheme convergence tolerance & 1e-5 \\ \hline
\begin{tabular}[c]{@{}l@{}}Maximum number of \\ staggered scheme iterations\end{tabular} & 20 \\ \hline
Newton solver tolerance & 1e-6 \\ \hline
Number of load increments & 20 \\ \hline
Number of elements along x & 128 \\ \hline
Number of elements along y & 64 \\ \hline
\end{tabular}
\end{table}

\FloatBarrier

\end{appendix}

\end{document}